\documentclass{article} 
\usepackage{iclr2027_conference,times}

\usepackage{amsmath,amsfonts,bm}

\def\eqref#1{equation~\ref{#1}}

\def\1{\bm{1}}

\DeclareMathAlphabet{\mathsfit}{\encodingdefault}{\sfdefault}{m}{sl}
\SetMathAlphabet{\mathsfit}{bold}{\encodingdefault}{\sfdefault}{bx}{n}

\usepackage{hyperref}
\usepackage{url}
\usepackage{booktabs}
\usepackage{graphicx}
\usepackage{float}

\usepackage[table]{xcolor}
\usepackage{colortbl}
\usepackage{amsmath,amssymb}
\usepackage{makecell}
\usepackage{caption}
\usepackage{tabularx}
\usepackage{array}

\newcolumntype{Y}{>{\centering\arraybackslash}X}

\newcommand{\scorepair}[2]{%
  \makebox[2.35em][r]{#1}%
  \makebox[0.85em][c]{/}%
  \makebox[2.35em][l]{#2}%
}

\usepackage{xcolor}
\usepackage{colortbl,booktabs,array,caption,float,ragged2e,tcolorbox}

\definecolor{gTwoRPerception}{HTML}{92500B} 
\definecolor{gTwoRSpatial}{HTML}{176263}   
\definecolor{gTwoRConstraint}{HTML}{465681}
\definecolor{gTwoRPlanning}{HTML}{75072D}  

\makeatletter

\def\gTwoR@registercolors#1#2{%
  \@for\gTwoR@task:=#2\do{%
    \expandafter\def
    \csname gTwoR@color@\gTwoR@task\endcsname{#1}%
  }%
}

\gTwoR@registercolors{gTwoRPerception}{Missing Tiles,Hidden Object,Dot Shape,Color Swap}
\gTwoR@registercolors{gTwoRSpatial}{Cube Net,Cylinder Coloring,Block Counting,Rolling Cube}
\gTwoR@registercolors{gTwoRConstraint}{Hexakai,Tapa,Battleship,Block Island,Pipe}
\gTwoR@registercolors{gTwoRPlanning}{Laser,KeyPath Quest,Growing Trail,Mancala,Rush Exit}

\providecommand{\gTwoRCompactCase}[5]{}
\renewcommand{\gTwoRCompactCase}[5]{%
  \begingroup

  \ifcsname gTwoR@color@#1\endcsname
    \edef\gTwoR@selectedcolor{%
      \csname gTwoR@color@#1\endcsname
    }%
  \else
    \def\gTwoR@selectedcolor{gTwoRPlanning}%
  \fi

  \begin{tcolorbox}[
    title={#1\hfill{\normalfont\small\##2}},
    colframe=\gTwoR@selectedcolor,
    colback=\gTwoR@selectedcolor!2!white,
    colbacktitle=\gTwoR@selectedcolor,
    coltitle=white,
    fonttitle=\small\bfseries,
    fontupper=\small,
    boxrule=0.5pt,
    arc=1mm,
    boxsep=0.7mm,
    left=1.5mm,
    right=1.5mm,
    top=1mm,
    bottom=1mm,
    before skip=6pt,
    after skip=6pt,
    before upper={%
      \setlength{\parindent}{0pt}%
      \setlength{\parskip}{1.5pt}%
      \raggedright
    }
  ]
    \textbf{Task.} #3\par
    \textbf{Question.} #4\par
    \textbf{Answer.} {\ttfamily #5}\par
  \end{tcolorbox}%

  \endgroup
}

\makeatother

\definecolor{groupblue}{RGB}{235,245,255}

\usepackage{tcolorbox}
\usepackage{longtable,ragged2e}

\definecolor{gTwoRCaseAccent}{HTML}{75072D}

\title{Game2Real-Bench: Do Toy Games Reveal and Transfer to Real-World Visual Reasoning?}

\author{%
\begin{minipage}[t]{\dimexpr\textwidth-2\tabcolsep\relax}
\centering
\normalfont\normalsize
\textbf{Qingan Liu}$^{1*}$,
\textbf{Juntong Feng}$^{1*}$,
\textbf{Yue Zhu}$^{1}$,
\textbf{Yujie Cheng}$^{1}$,
\textbf{Xinzhe Han}$^{1}$ \\
\textbf{Shuai Hao}$^{1}$,
\textbf{Yunzhi Zhuge}$^{1}$,
\textbf{Haiwen Diao}$^{2\dagger}$,
\textbf{Huchuan Lu}$^{1}$\\[0.3em]
$^{1}$Dalian University of Technology,
$^{2}$Nanyang Technological University
\endgraf
\end{minipage}%
}

\iclrfinalcopy

\begin{document}

\maketitle

\pagestyle{fancy}
\fancyhf{}
\fancyfoot[C]{\thepage}
\renewcommand{\headrulewidth}{0pt}
\renewcommand{\footrulewidth}{0pt}
\fancypagestyle{preprintfirstpage}{%
    \fancyhf{}
    \fancyfoot[L]{\footnotesize Preprint.}
    \renewcommand{\headrulewidth}{0pt}
    \renewcommand{\footrulewidth}{0pt}
}
\thispagestyle{preprintfirstpage}

\ificlrfinal
\begingroup
\renewcommand{\thefootnote}{}
\footnotetext[0]{%
$^{*}$Equal contribution.
$^{\dagger}$Correspondence to
\texttt{haiwen.diao@ntu.edu.sg}
}
\endgroup
\fi

\begin{abstract}

Visual reasoning is an important capability for vision-language models (VLMs)
to understand visual evidence, spatial structure, constraints, and state
changes. Existing evaluations mainly rely on either real-world data or
procedurally generated tasks. Real-world data preserve the complexity of
natural scenes, while procedural games provide controllable generation and
verifiable solutions. However, it remains unclear whether the abilities
measured in toy games correspond to those required in real-world visual
reasoning, and whether game examples can provide useful signals for solving
real-world tasks.
We introduce \textbf{Game2Real-Bench}, a benchmark for studying these
connections across four capability groups, including fine-grained perception,
spatial reasoning, constraint reasoning, and sequential planning. On the game
side, we construct 1,000 independently generated samples from executable task
specifications and verify their answers programmatically. On the real-world
side, we curate and construct 400 samples from diverse real-world sources.
We connect the two settings according to shared reasoning requirements,
including visual evidence, spatial operations, constraints, and state
transitions, rather than visual similarity.
We evaluate 25 representative VLMs across both settings. Performance on game
and real-world tasks shows consistent cross-setting trends, with a Pearson
correlation of 0.935 and a Spearman correlation of 0.935. We further examine
game-to-real transfer through one-shot in-context learning on four models.
The effect varies across capabilities and models. Constraint reasoning shows
positive gains for several models, while fine-grained perception consistently
declines with game demonstrations. These results show that toy games can reflect
some reasoning patterns observed in real-world tasks and can provide useful
demonstrations in selected settings.
To further support model training, we construct
\textbf{Game2Real-Dataset}, comprising nearly 24K game samples across
18 tasks and 65 QA families. These samples are generated from executable
task specifications and verified programmatically. Our benchmark, dataset,
and code are available at
\url{https://github.com/QingAnLiu/Game2Real-Bench}.

\end{abstract}

\section{Introduction}

Recent advances in vision-language models (VLMs) have led to substantial
progress in multimodal understanding and
reasoning~\citep{bai2025qwen3,wang2025internvl3,diao2026pixels,diao2026sensenova}.
Beyond recognizing what is present in an image, visual reasoning requires
models to understand how objects are related, how constraints interact,
and how actions change a scene. These demands arise in everyday tasks,
such as interpreting an assembly step, as well as in toy games with
explicit rules and controllable states. Games are particularly appealing
because their tasks can be generated at scale and their answers verified
programmatically~\citep{tong2026game,xu2026vbvr}. Yet their simplified
visual environments raise a central question: \textit{Do toy games reveal
VLMs' real-world visual reasoning capabilities, and can solved game
examples support transfer to real-world tasks?}

Existing benchmarks characterize different aspects of visual reasoning.
MMMU and MME-RealWorld cover knowledge-intensive problems and real-world
scenarios~\citep{yue2024mmmu,zhang2025mme}, while BLINK and BabyVision
reveal weaknesses in foundational visual skills despite strong high-level
reasoning performance~\citep{fu2024blink,chen2026babyvision}. Recent work
shows that procedural tasks can support reasoning beyond their original
settings: Game-RL improves VLM reasoning through game-based
training~\citep{tong2026game}, and VBVR-Pro demonstrates transfer to
external reasoning benchmarks for visual generation
models~\citep{xu2026vbvr}. These findings highlight the potential of
procedural training, but leave open how game-based assessment relates
to real-world reasoning across models and capabilities. A shared
evaluation linking tasks by their reasoning requirements is needed to
examine this relationship and the usefulness of game examples.

To this end, we introduce \textbf{Game2Real-Bench}, a benchmark linking
toy games and real-world visual reasoning across four capabilities:
\textit{Fine-grained Perception}, \textit{Spatial Reasoning},
\textit{Constraint Reasoning}, and \textit{Sequential Planning}.
We construct the game evaluation using executable generators and
task-specific verifiers, and curate real-world tasks according to
shared reasoning requirements rather than visual resemblance
(Figure~\ref{fig:pipeline}). The resulting benchmark combines 1,000
independently generated and programmatically verified toy-game samples
with 400 real-world samples derived from images, videos, and multi-view
observations. The same pipeline produces \textbf{Game2Real-Dataset},
a separate collection of nearly 24K verified samples spanning
18 toy-game tasks and 65 question-answer (QA) families, providing
a scalable resource for visual reasoning evaluation and training.

Our evaluation of 25 representative VLMs reveals a strong correspondence
between game and real-world performance, with model-level Pearson and
Spearman correlations of $0.935$. Despite their simpler visual
environments, game tasks are generally harder and better distinguish
models. The benefits of game examples are more selective. In one-shot
in-context learning with four models, randomly selected game
demonstrations improve constraint reasoning for several models but
lower fine-grained perception accuracy for all four, with a slight
decline in average overall accuracy. These findings suggest that toy
games are informative probes of real-world visual reasoning, while
their usefulness as demonstrations depends on the capability and model.
We further analyze thinking mode and recurring failures in spatial
representation, relational consistency, and state updating.

Our main contributions are summarized as follows:
\begin{itemize}
    \item We introduce \textbf{Game2Real-Bench}, a 1,400-sample benchmark
    that aligns toy-game and real-world tasks through shared reasoning
    requirements across four capabilities, enabling analysis of
    cross-setting performance correspondence and game-to-real transfer.

    \item We construct \textbf{Game2Real-Dataset}, comprising nearly 24K
    programmatically verified samples across 18 toy-game tasks and
    65 QA families, together with executable generators and task-specific
    verifiers to support future research on scalable visual reasoning.

    \item We evaluate 25 VLMs, revealing strong game--real performance
    correspondence and selective one-shot transfer, with analyses of
    thinking mode and reasoning failures.
\end{itemize}

\section{Related Work}

\subsection{Games as Testbeds for Visual Reasoning}

Recent work has increasingly used games and procedural environments as testbeds
for multimodal reasoning, benefiting from their explicit rules, controllable
states, and verifiable outcomes. In interactive settings, benchmarks such as
BALROG, LMGame-Bench, VideoGameBench, and GameWorld use diverse games to evaluate
perception, planning, memory, and decision making in LLM and VLM agents
\citep{paglieri2025balrog,hu2026lmgame,zhang2025videogamebench,ouyang2026gameworld}.
Beyond evaluation, games have also been used to construct reasoning tasks and
training supervision. Game-RL synthesizes verifiable multimodal reasoning
problems from game mechanics, while Play to Generalize investigates whether
reasoning abilities acquired through gameplay transfer to broader multimodal
tasks \citep{tong2026game,xie2026play}. VBVR-Pro further scales procedural
visual reasoning to a broad collection of automatically generated tasks with
task-specific verification \citep{xu2026vbvr}. Together, these studies establish
games and procedural environments as effective testbeds for multimodal reasoning.
However, their relationship to real-world reasoning is typically assessed
indirectly, either through in-game performance or aggregate results on external
benchmarks. Game2Real-Bench makes this relationship explicit by organizing
procedurally generated game tasks and real-world samples around shared
reasoning requirements, enabling analysis of cross-setting
performance correspondence and game-to-real transfer at the capability level.

\subsection{Benchmarks for Visual Reasoning}

Recent advances in encoder-based and encoder-free
VLMs~\citep{diao2024unveiling,diao2025evev2} have improved visual
perception, understanding, and multimodal reasoning.
To evaluate these capabilities, a wide range of benchmarks have been proposed. General-purpose
benchmarks such as MMMU and MME-RealWorld assess multimodal reasoning across
knowledge-intensive and complex real-world scenarios
\citep{yue2024mmmu,zhang2025mme}, while vision-centric benchmarks such as
BLINK, MMVP, and BabyVision focus more directly on visual perception and
reasoning with reduced reliance on external knowledge
\citep{fu2024blink,tong2024eyes,chen2026babyvision}. Other benchmarks study
more structured forms of visual reasoning, including 3D understanding,
multi-view reasoning, spatial relations, and logical constraints
\citep{jia2026omnispatial,huang2026spatial,xie2026spatialqa}.
Recent work has also explored synthetic and procedural tasks for controlled
visual reasoning. CLEVR uses synthetic scenes to study compositional visual
reasoning \citep{johnson2017clevr}, while later efforts such as Game-RL, VBVR,
and VBVR-Pro extend procedural generation to broader reasoning tasks, games,
and training settings
\citep{tong2026game,wang2026very,xu2026vbvr}. These settings enable precise
control over task structure and reliable solution generation at scale.
However, it is still not well understood whether the capabilities reflected by
procedurally generated games correspond to those required by real-world visual
reasoning. Game2Real-Bench studies this relationship by combining game and
real-world visual reasoning tasks under a shared capability framework,
allowing us to examine both cross-setting performance consistency and
game-to-real transfer.

\begin{figure}[t]
    \centering
    \includegraphics[width=\linewidth]{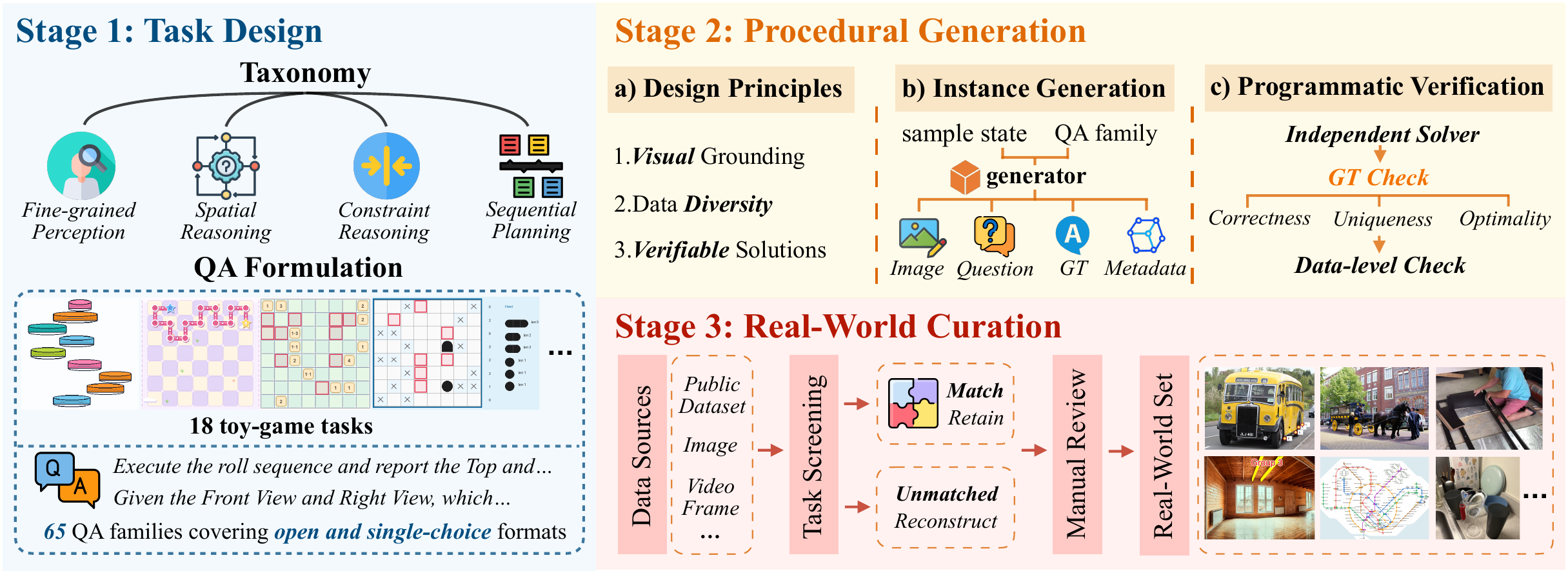}
    \vspace{-5mm}
    \caption{\textbf{Multi-stage construction pipeline of Game2Real-Bench.}
    It comprises capability-guided task design, procedural generation and verification,
    and mechanism-aligned real-world curation.}
    \label{fig:pipeline}
\end{figure}

\section{Game2Real-Bench}

We construct Game2Real-Bench in three stages, as illustrated in
Figure~\ref{fig:pipeline}. In Stage 1, we organize the task space around four
capability groups and design 18 toy-game tasks with 65 QA families. In Stage 2,
we develop executable generators and task-specific verifiers to construct
\textbf{Game2Real-Dataset} with 23,893 verified game samples, while independently
generating 1,000 samples for the toy-game evaluation set. In Stage 3, we curate
400 real-world samples that require reasoning operations corresponding to those
exercised by the toy games. The outputs of Stages 2 and 3 together constitute
the complete 1,400-sample Game2Real-Bench. We describe each stage in detail
below.

\subsection{Task Design}

\textbf{Taxonomy.}
We organize Game2Real-Bench around four complementary capabilities in visual
reasoning and design 18 toy-game tasks accordingly.
\textbf{\textit{Fine-grained Perception}} requires models to identify
task-relevant visual evidence from subtle local differences, correspondences,
and partial observations.
\textbf{\textit{Spatial Reasoning}} requires models to infer and transform
spatial configurations involving relations, orientation, viewpoint, geometry,
and 3D structure.
\textbf{\textit{Constraint Reasoning}} requires models to integrate multiple
interdependent relations and constraints to derive a globally consistent
solution.
\textbf{\textit{Sequential Planning}} requires models to reason over state
transitions and select state-dependent actions toward a target goal.
Individual tasks may involve multiple capabilities, and each task is assigned
to the group that best reflects its primary reasoning demand.

\textbf{Task and QA Formulation.}
In Game2Real-Bench, a toy-game task defines a structured visual reasoning
problem with explicit rules and solution criteria, while its QA families
specify the reasoning targets and output formats. These QA families
cover fine-grained visual evidence, spatial relations and transformations,
constraint-consistent configurations, and action selection over evolving
states. Across the 18 tasks, we define 65 QA families in open-ended and
multiple-choice formats. A sample is generated from a task configuration and
contains one or more visual observations, a question, and a verified answer.
The corresponding QA family further specifies any task-specific notation,
coordinate system, and output convention, ensuring consistent evaluation
across generated samples.

\begin{figure}[t]
    \centering
    \includegraphics[
        width=\linewidth,
        trim={0 5mm 0 0},
        clip
    ]{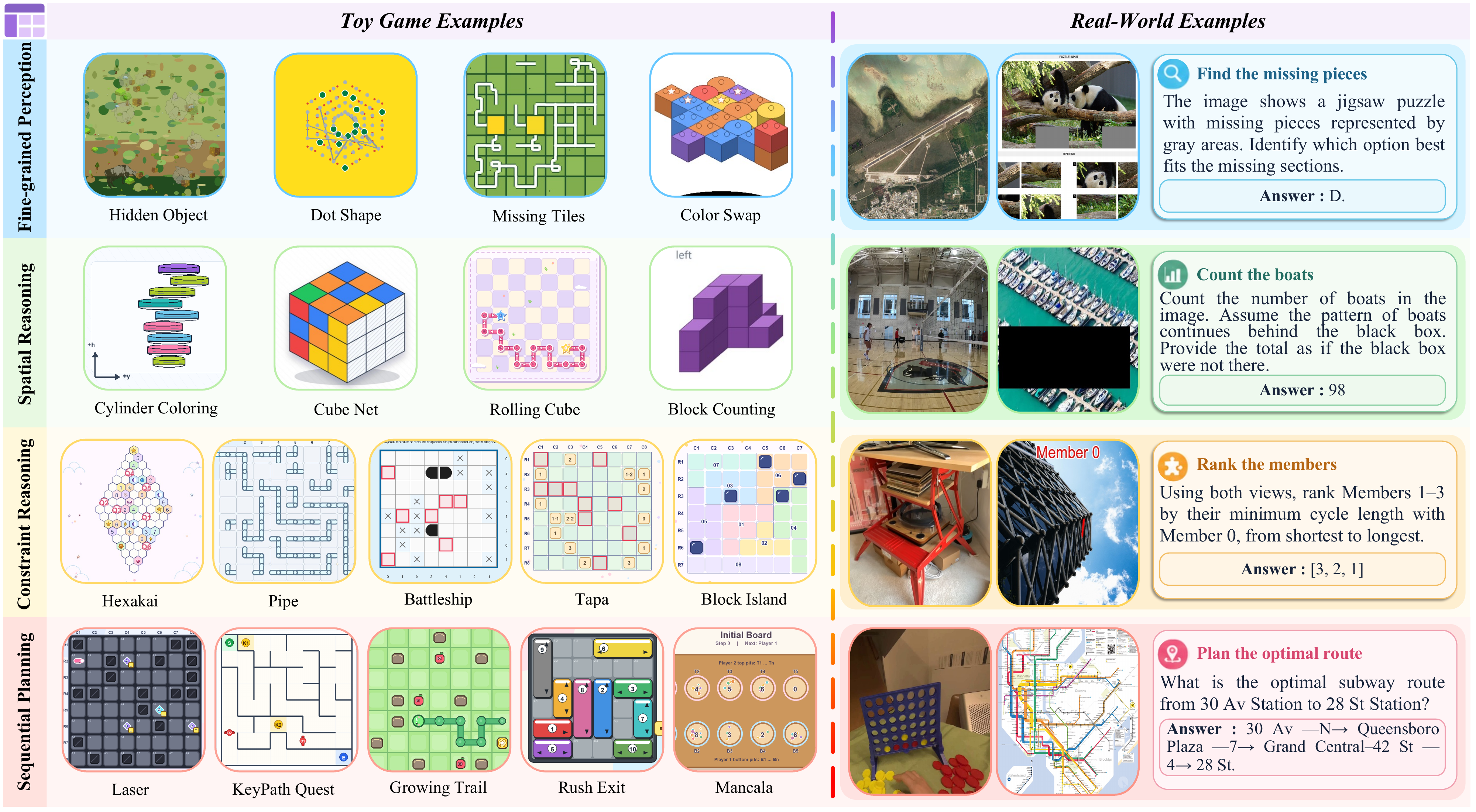}
    \vspace{-5mm}
    \caption{\textbf{Capability-aligned tasks in Game2Real-Bench.}
    Toy-game and real-world tasks span fine-grained perception, spatial
    reasoning, constraint reasoning, and sequential planning. We present
    representative QA samples illustrating shared reasoning requirements
    across the two settings.}
    \label{fig:category}
\end{figure}

\subsection{Procedural Generation}

\textbf{Design Principles.}
Our procedural generation follows three principles.
\textbf{\textit{Visual Grounding.}} Each question is designed so that its answer is
determined by the visual content rather than prompt wording. The text specifies
the task and rules, while the image provides the instance-specific evidence
required to solve it.
\textbf{\textit{Diversity.}} We vary problem structures, spatial layouts, and visual
presentations across generated instances while preserving the underlying task
rules and reasoning objective, producing diverse instances of the same capability.
\textbf{\textit{Verifiability.}} Each problem follows explicit executable rules, allowing
the correct solution to be derived and checked programmatically.

\textbf{Instance Generation.}
For each task, we implement a task-specific generator that samples valid problem
states from predefined configuration ranges. Given a sampled state and QA family,
the generator jointly produces the visual observations, question, and
ground-truth answer. Structural parameters control factors such as topology,
object configurations, constraints, paths, and state transitions, while
presentation parameters determine how each state is visually encoded and
rendered. We also retain the structured metadata required for verification,
analysis, and de-duplication.

\textbf{Programmatic Verification.}
Each task is paired with a deterministic solver or verifier tailored to its
solution structure. Verification may involve constraint solving, geometric
transformation, state simulation, graph search, or transition replay.
Given the sampled state, the verifier recomputes the requested result or
checks the proposed solution against executable task rules. For tasks requiring uniqueness,
alternative valid solutions are explicitly checked; when multiple solutions
are permitted, all admissible answers are retained. When optimality is
required, the answer is checked against the solver-derived optimum.
Verified samples additionally pass shared checks for record completeness,
image validity, answer-space consistency, multi-image alignment, and
required metadata.

\subsection{Real-World Curation}

We construct the real-world evaluation set by curating samples from existing
visual reasoning benchmarks and public datasets. Candidate samples are drawn
from sources containing real images, videos, multi-view observations, and
structured annotations
\citep{
fu2024blink,
wu2026visual,
ji2025visualtrans,
yeh2026seeing,
yang2026thinking,
feng2026reasonmap,
pothiraj2025capture,
wang2021ocid,xie2026spatialqa}. We select samples whose reasoning
requirements align with those of the toy-game tasks, including fine-grained
visual evidence, spatial structure, relational constraints, and state changes.
For benchmark tasks that already meet the target reasoning requirements, we
retain their original questions, answers, and evaluation protocols. Otherwise,
we construct new evaluation samples from the source media and annotations while
preserving the same reasoning requirements. For example, for IKEA Manuals at
Work \citep{liu2024ikea}, we align assembly instructions with real assembly
videos and construct multi-image questions from visually distinguishable states
before and after an assembly step. All samples in the final set undergo automated
consistency checks and manual review, and samples with ambiguous answers or
insufficient visual evidence are removed.

\begin{table*}[t]
\centering
\caption{
Scores (\%) on Game2Real-Bench, with a 1,000-sample Toy Game set
and a 400-sample Real-World set.
For each capability, results are reported as Toy Game / Real-World.
Toy-game scores are accuracies; real-world scores incorporate
SpatiaLQA F1 where applicable.
Overall denotes the mean per-sample score over all 1,400 samples.
Open-weight models are grouped by total parameter count as large
(27--35B), medium (7--12B), and small (2--4B).
For MoE models, active parameter counts are retained in the model names.
The best Toy Game and Real-World results within each group are
highlighted in \textbf{bold}; models are ordered by Overall score.
}
\label{tab:main_results}
\vspace{0.4em}

\small
\setlength{\tabcolsep}{5.5pt}
\renewcommand{\arraystretch}{1.08}

\begin{tabularx}{\textwidth}{
@{}
>{\raggedright\arraybackslash}m{0.28\textwidth}|
Y|Y|Y|Y|Y
@{}
}

\toprule

\makecell[l]{\textbf{Model}}
&
\makecell[c]{\textbf{Fine-grained}\\\textbf{Perception}}
&
\makecell[c]{\textbf{Spatial}\\\textbf{Reasoning}}
&
\makecell[c]{\textbf{Constraint}\\\textbf{Reasoning}}
&
\makecell[c]{\textbf{Sequential}\\\textbf{Planning}}
&
\makecell[c]{\textbf{Overall}}
\\

\midrule


\rowcolor{groupblue}
\multicolumn{6}{l}{
\textbf{$\blacktriangledown$ Proprietary Models}
}
\\

GPT-5.6-Sol
& \scorepair{\textbf{53.9}}{57.0}
& \scorepair{\textbf{69.8}}{49.0}
& \scorepair{\textbf{81.3}}{\textbf{63.9}}
& \scorepair{\textbf{81.4}}{68.0}
& \textbf{69.8}
\\

Gemini-3.1-Pro-Preview
& \scorepair{36.0}{\textbf{66.0}}
& \scorepair{59.0}{\textbf{55.0}}
& \scorepair{56.7}{51.1}
& \scorepair{75.2}{\textbf{73.0}}
& 59.6
\\

Claude-Opus-4.8-thinking
& \scorepair{22.5}{50.0}
& \scorepair{50.5}{44.0}
& \scorepair{62.0}{46.8}
& \scorepair{70.8}{61.0}
& 53.7
\\

\midrule


\rowcolor{groupblue}
\multicolumn{6}{l}{
\textbf{$\blacktriangledown$ Large Open-Weight Models}
}
\\

Qwen3.8-27B
& \scorepair{\textbf{37.6}}{\textbf{67.0}}
& \scorepair{\textbf{61.3}}{\textbf{55.0}}
& \scorepair{39.3}{38.8}
& \scorepair{\textbf{61.0}}{\textbf{72.0}}
& \textbf{52.6}
\\

Qwen3.5-35B-A3B
& \scorepair{31.5}{57.0}
& \scorepair{42.3}{53.0}
& \scorepair{33.4}{\textbf{48.2}}
& \scorepair{51.5}{67.0}
& 45.0
\\

Qwen3.5-27B
& \scorepair{28.6}{54.0}
& \scorepair{48.6}{51.0}
& \scorepair{34.1}{36.6}
& \scorepair{54.6}{63.0}
& 44.9
\\

Gemma4-31B-it
& \scorepair{24.2}{57.0}
& \scorepair{25.7}{38.0}
& \scorepair{\textbf{42.3}}{46.5}
& \scorepair{40.0}{50.0}
& 38.5
\\

Qwen3-VL-30B-A3B-Instruct
& \scorepair{10.7}{42.0}
& \scorepair{17.6}{26.0}
& \scorepair{16.4}{26.1}
& \scorepair{19.3}{46.0}
& 21.8
\\

InternVL3.5-30B-A3B
& \scorepair{15.7}{37.0}
& \scorepair{16.7}{26.0}
& \scorepair{17.1}{27.6}
& \scorepair{15.9}{39.0}
& 21.0
\\

\midrule


\rowcolor{groupblue}
\multicolumn{6}{l}{
\textbf{$\blacktriangledown$ Medium Open-Weight Models}
}
\\

Qwen3.5-9B
& \scorepair{\textbf{27.0}}{\textbf{51.0}}
& \scorepair{\textbf{25.2}}{\textbf{52.0}}
& \scorepair{24.3}{24.3}
& \scorepair{\textbf{47.8}}{\textbf{54.0}}
& \textbf{35.7}
\\

Step3-VL-10B
& \scorepair{12.9}{34.0}
& \scorepair{13.5}{31.0}
& \scorepair{21.3}{26.4}
& \scorepair{29.8}{41.0}
& 24.2
\\

Gemma4-12B-it
& \scorepair{16.9}{48.0}
& \scorepair{15.8}{27.0}
& \scorepair{\textbf{25.6}}{22.3}
& \scorepair{21.4}{24.0}
& 23.4
\\

GLM-4.6V-Flash
& \scorepair{16.3}{31.0}
& \scorepair{15.8}{34.0}
& \scorepair{17.1}{\textbf{30.0}}
& \scorepair{19.0}{44.0}
& 22.2
\\

Ovis2.5-9B
& \scorepair{18.5}{26.0}
& \scorepair{15.3}{23.0}
& \scorepair{16.4}{28.9}
& \scorepair{18.0}{35.0}
& 20.2
\\

NEO-ov-9B
& \scorepair{13.5}{32.0}
& \scorepair{20.3}{26.0}
& \scorepair{14.1}{26.7}
& \scorepair{13.9}{27.0}
& 18.9
\\

InternVL3.5-8B
& \scorepair{11.2}{29.0}
& \scorepair{17.6}{25.0}
& \scorepair{13.4}{26.1}
& \scorepair{17.3}{32.0}
& 18.8
\\

SenseNova-U1-8B-MoT
& \scorepair{13.5}{23.0}
& \scorepair{16.7}{33.0}
& \scorepair{12.5}{15.3}
& \scorepair{18.0}{36.0}
& 18.5
\\

Qwen3-VL-8B-Instruct
& \scorepair{13.5}{33.0}
& \scorepair{13.1}{21.0}
& \scorepair{12.8}{25.5}
& \scorepair{15.9}{33.0}
& 18.0
\\

MiMo-VL-7B-RL
& \scorepair{13.5}{28.0}
& \scorepair{13.5}{15.0}
& \scorepair{11.8}{15.1}
& \scorepair{18.3}{39.0}
& 17.2
\\

NEO-9B-SFT
& \scorepair{13.5}{32.0}
& \scorepair{12.6}{18.0}
& \scorepair{11.5}{21.6}
& \scorepair{16.6}{30.0}
& 17.0
\\

\midrule


\rowcolor{groupblue}
\multicolumn{6}{l}{
\textbf{$\blacktriangledown$ Small Open-Weight Models}
}
\\

Qwen3.5-4B
& \scorepair{\textbf{20.8}}{\textbf{59.0}}
& \scorepair{\textbf{15.8}}{\textbf{38.0}}
& \scorepair{\textbf{23.6}}{\textbf{28.9}}
& \scorepair{\textbf{40.7}}{\textbf{54.0}}
& \textbf{31.7}
\\

Qwen3.5-2B
& \scorepair{11.8}{41.0}
& \scorepair{8.6}{23.0}
& \scorepair{18.7}{16.5}
& \scorepair{15.2}{47.0}
& 19.3
\\

NEO-2B-SFT
& \scorepair{9.0}{28.0}
& \scorepair{9.0}{21.0}
& \scorepair{9.5}{15.1}
& \scorepair{10.2}{25.0}
& 13.2
\\

Qwen3-VL-2B-Instruct
& \scorepair{6.7}{26.0}
& \scorepair{10.4}{23.0}
& \scorepair{8.5}{23.8}
& \scorepair{6.1}{25.0}
& 12.6
\\

InternVL3.5-2B
& \scorepair{10.7}{26.0}
& \scorepair{11.7}{17.0}
& \scorepair{7.9}{14.8}
& \scorepair{7.1}{27.0}
& 12.5
\\

\bottomrule
\end{tabularx}

\end{table*}

\subsection{Benchmark Statistics}

\textbf{Game2Real-Dataset.}
Game2Real-Dataset contains 23,893 verified game samples across four capability
groups, 18 toy-game tasks, and 65 QA families. It comprises 20,701 open-ended
and 3,192 single-choice samples, with both single-image and multi-image inputs
and 36,441 images in total.

\textbf{Game2Real-Bench.}
Game2Real-Bench contains 1,400 evaluation samples, including 1,000 toy-game
samples and 400 real-world samples. The toy-game set is generated independently
from Game2Real-Dataset using disjoint random seeds and distinct layout
configurations. We further use semantic fingerprints and image hashes to remove
potential overlap. The set includes 600 open-ended and 400 single-choice
questions. For the real-world set, we curate 400 samples to provide broad coverage across
the four capability groups and their corresponding reasoning requirements. The Game2Real-Bench
composition and representative examples are shown in
Figure~\ref{fig:category}.

\section{Experiments}

\subsection{Setup}

\textbf{Models.}
We evaluate 25 vision-language models, including 22 open-weight models and
three proprietary frontier systems. The open-weight models span 2B to 35B
total parameters and cover representative families including
Qwen3-VL~\citep{bai2025qwen3},
Qwen3.5~\citep{qwen3.5}, Qwen3.8~\citep{qwen2026qwen38},
InternVL3.5~\citep{wang2025internvl3}, Gemma4~\citep{team2026gemma},
GLM-4V~\citep{hong2025glm}, MiMo-VL~\citep{yue2025mimo},
Ovis2.5~\citep{lu2025ovis2}, NEO~\citep{diao2026pixels}, NEO-ov~\citep{diao2026pixelsov}, SenseNova-U1~\citep{diao2026sensenova}, and Step3-VL~\citep{huang2026step3}.
We additionally evaluate three proprietary frontier models GPT-5.6-Sol~\citep{openai2026gpt56},
Gemini-3.1-Pro-Preview~\citep{google2026gemini31pro}, and
Claude-Opus-4.8~\citep{anthropic2026opus48}.  Unless otherwise
specified, all models use default inference configurations.

\textbf{Evaluation Protocol.}
All models are evaluated zero-shot using the same task instructions, source
visual inputs, and questions. Open-weight models are evaluated through
VLMEvalKit~\citep{duan2024vlmevalkit} using their official checkpoints and
model-specific preprocessing, while proprietary models are accessed through
their official APIs. No external tools or task-specific solvers are used
during inference. For toy-game samples, model outputs are normalized
according to each QA family's answer schema and scored by deterministic
evaluators derived from the task rules. For real-world samples, we follow the original
evaluation protocol when a source benchmark provides a standard evaluator and
apply unified evaluation criteria to reconstructed samples. We report
capability-level performance and the overall average score over the
corresponding evaluation sets.


\subsection{Main Results}

Table~\ref{tab:main_results} presents the performance of 25 VLMs on the
toy-game and real-world evaluation sets.

\textbf{Current VLMs remain challenged on Game2Real-Bench.}
GPT-5.6-Sol achieves the highest overall score at $69.8\%$, while
Gemini-3.1-Pro-Preview and Claude-Opus-4.8 reach $59.6\%$ and
$53.7\%$, respectively. Among open-weight models, Qwen3.8-27B performs best
at $52.6\%$, while 16 of the 22 models score below $30\%$. These results
indicate that Game2Real-Bench remains difficult for current VLMs, with clear
room for improvement across both proprietary and open-weight models.

\textbf{Toy-game tasks are more challenging and provide clearer separation
between models.}
Most models achieve lower performance on the toy-game set than on the
real-world set, with the difference being particularly visible among
open-weight models. The toy-game evaluation also produces wider gaps between
stronger and weaker systems. For example, the best proprietary and open-weight
models differ by $16.3$, $39.0$, and $20.4$ points on fine-grained perception,
constraint reasoning, and sequential planning, respectively, while these gaps
are much smaller on the corresponding real-world tasks. This difference may
partly reflect the controlled nature of the toy-game tasks, where models must
follow explicit rules and produce solutions that satisfy precisely defined
states, constraints, and transitions. Real-world samples provide more familiar
visual content and generally yield a more compressed performance range. This 
contrast highlights the different roles of the two settings. Real-world tasks 
assess reasoning in natural visual contexts, while toy-game
tasks provide finer discrimination among current VLMs under more explicitly
specified reasoning requirements.

\textbf{Performance varies considerably across capabilities, with sequential
planning showing the most consistent strength.}
Sequential planning yields the strongest performance for many models across
both evaluation settings, and is the highest-scoring capability for 15 of the
25 models on the toy-game set and 17 on the real-world set. In contrast, the
relative difficulty of the other capabilities changes across settings.
Fine-grained perception produces the lowest average performance on the toy-game
set, but becomes one of the stronger capabilities on real-world tasks.
Constraint reasoning shows the opposite pattern more clearly among stronger
models, remaining difficult and preserving a visible gap between proprietary
and open-weight systems. These results suggest that capability performance is
not determined by the reasoning category alone, but also by how the underlying
reasoning requirements are instantiated in each visual setting.

\begin{figure}[t]
    \centering
    \includegraphics[width=\linewidth]{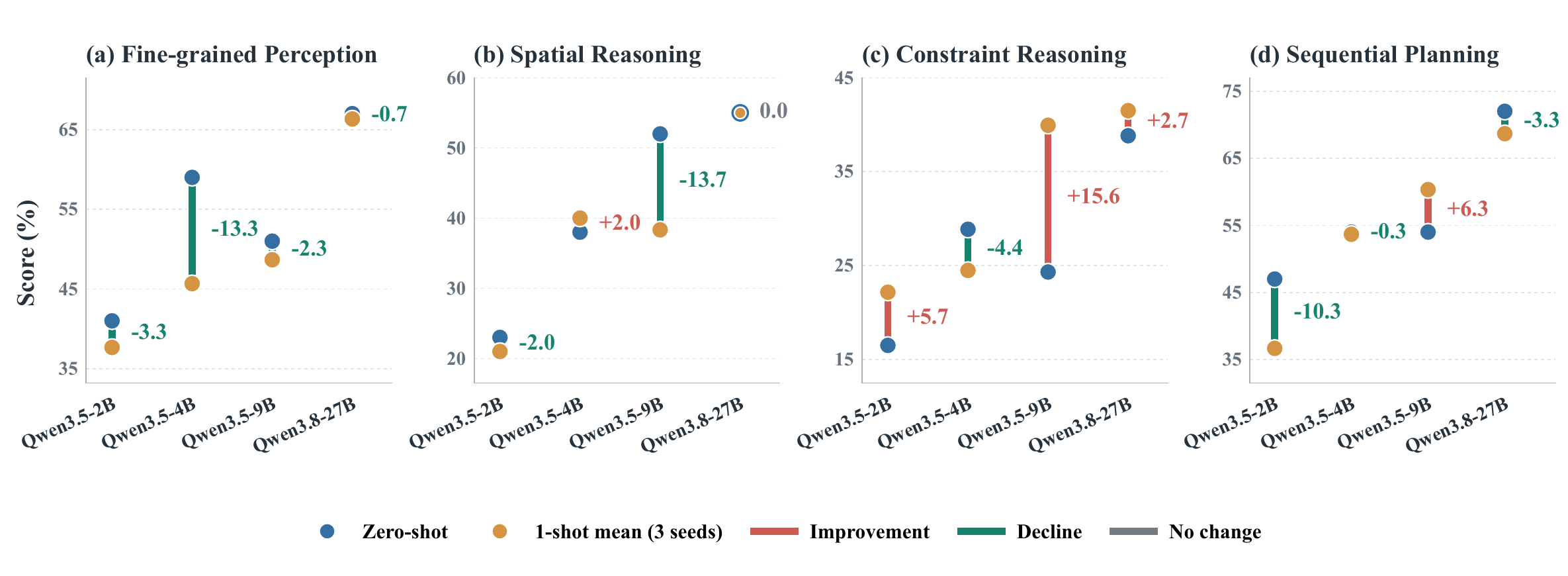}
    \vspace{-5mm}
    \caption{\textbf{Capability-wise game-to-real transfer with one-shot
    in-context learning.}
    Compared with zero-shot evaluation, random game demonstrations improve
    constraint reasoning on average but reduce fine-grained perception
    accuracy for all four models on the 400 real-world targets.}
    \label{fig:icl_transfer}
\end{figure}

\subsection{Fine-grained Analysis}


To better understand the relation between game tasks and real-world visual
reasoning, we analyze model-level consistency across the two evaluation
settings, game-to-real transfer through one-shot in-context learning, the
effect of thinking mode, and representative failure cases.

\textbf{Game-to-real transfer with one-shot in-context learning.}
We study whether solved game examples can transfer to real-world visual
reasoning through one-shot in-context learning. We evaluate four representative
models on the same 400 real-world targets used in the zero-shot setting. We
construct a pool of 168 verified samples from Game2Real-Dataset covering
14 game tasks, and randomly assign one game demonstration to each target while
balancing the four capabilities across the full set. We repeat the assignment
with three random seeds and use the same demonstration-target pairs for all
models. Each demonstration contains the game image and question followed only
by its ground-truth final answer. All other evaluation
settings remain unchanged.

The effect of game demonstrations varies clearly across both capabilities and
models, as shown in Figure~\ref{fig:icl_transfer}. Constraint reasoning improves
by $4.89$ points on average, with Qwen3.5-9B increasing from $24.32\%$ to
$39.94\%$. These tasks require models to abstract visual observations into
rules, relations, and valid configurations, so a solved game example may help
establish a structured problem-solving context even when the demonstration
comes from a different task. Fine-grained perception shows the opposite
pattern, decreasing by $4.92$ points on average and declining for all four
models. These tasks already achieve relatively high zero-shot performance and
depend mainly on target-specific visual evidence, for which an unrelated
demonstration provides little additional information and may introduce
irrelevant context. The model-level results are also non-uniform. Qwen3.5-9B
is the only model with an overall gain of $1.49$ points, while Qwen3.8-27B
changes by only $-0.33$ points, suggesting limited sensitivity to a single
random demonstration. The 2B and 4B models decrease overall, which may reflect
greater difficulty in separating useful contextual cues from unrelated task
information. Overall accuracy decreases by $1.34$ points on average, indicating
that game-to-real transfer is selective and depends on both the reasoning
requirement and the model.

\begin{figure*}[t]
    \centering

    \begin{minipage}[t]{0.49\textwidth}
        \centering
        \includegraphics[width=\linewidth]
        {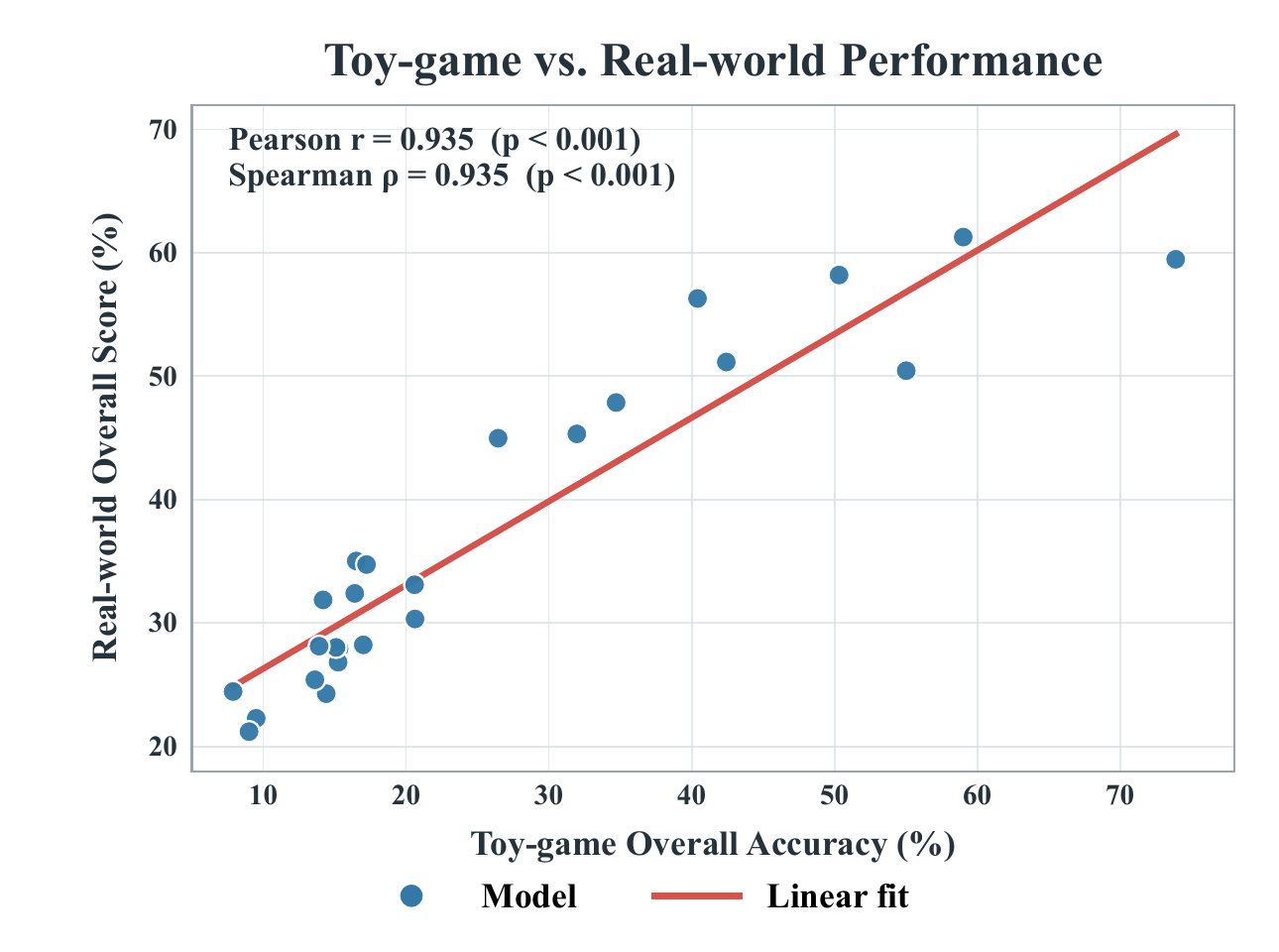}
        \vspace{-3mm}

        \caption{\textbf{Game-to-real performance correlation.}
        Model performance across the game and real-world evaluations is strongly correlated.}
        
        \label{fig:game_real_correlation}
    \end{minipage}
    \hfill
    \begin{minipage}[t]{0.49\textwidth}
        \centering
        \includegraphics[width=\linewidth]
        {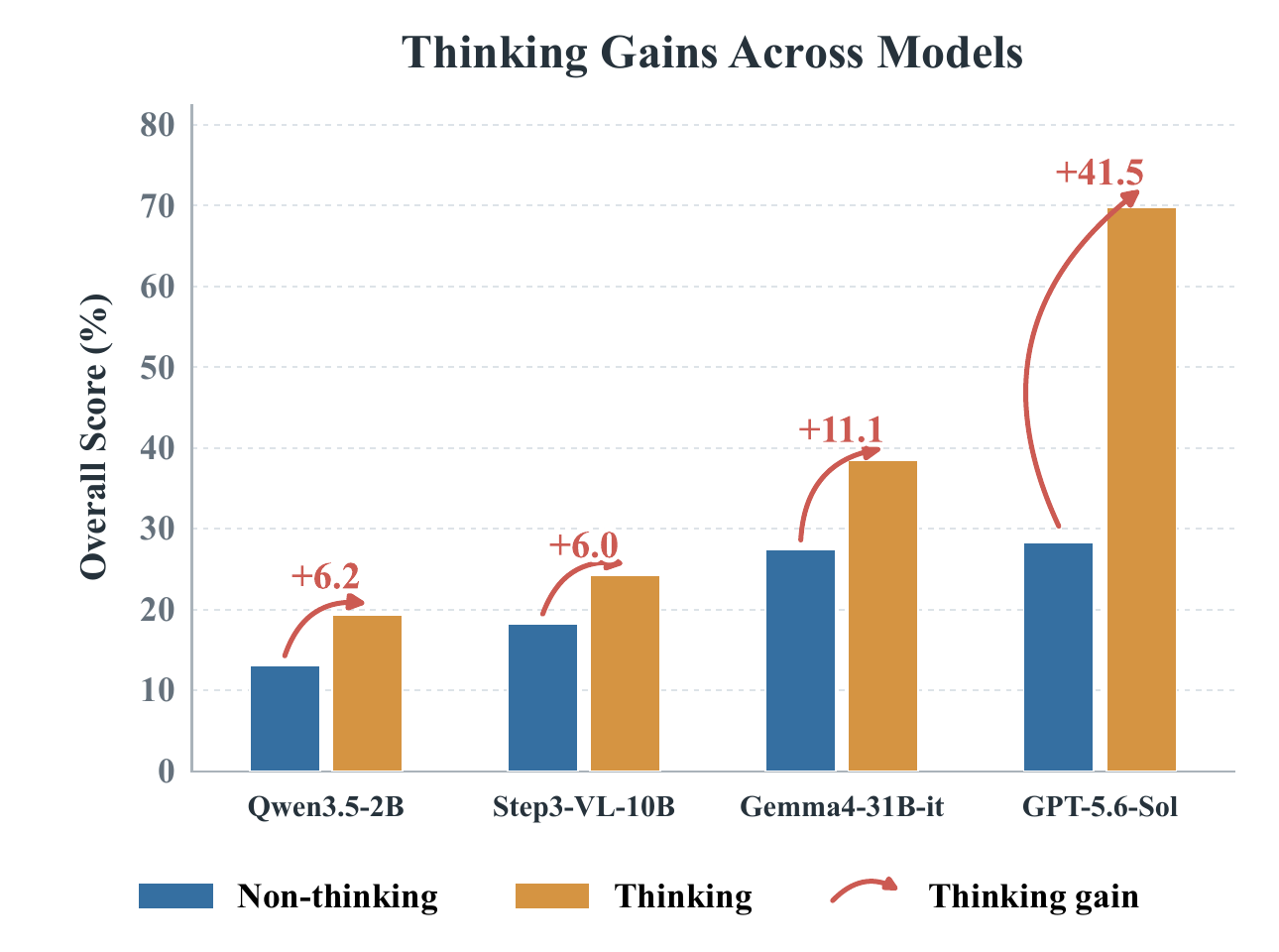}
        \vspace{-3mm}

        \caption{\textbf{Thinking gains across model scales.}
        Thinking improves overall score across all four models, with larger models benefiting more.}
        \label{fig:thinking_analysis}
    \end{minipage}

    \vspace{-4mm}
\end{figure*}

\textbf{Consistency between game and real-world evaluation.}
We examine whether the controlled game evaluation preserves the model-level
performance differences observed on real-world visual reasoning tasks. This analysis covers all 25 models, with each model treated as one observation. For each model, we pair
its overall score on the 1,000 game samples with its independently computed
overall score on the 400 real-world samples. We then compute Pearson and
Spearman correlations across the 25 model pairs. Pearson correlation measures
the linear association between model scores in the two settings, while
Spearman correlation measures whether their relative rankings are preserved.

We obtain a Pearson correlation of $r=0.935$ and a Spearman correlation of
$\rho=0.935$, as shown in Figure~\ref{fig:game_real_correlation}. 
Both metrics indicate that the two evaluations
preserve similar model-level performance patterns. Models with higher scores on
the game evaluation generally also achieve higher scores on the real-world
evaluation, and their relative ordering remains largely consistent across the
two settings. This agreement suggests that, despite differences in visual
content and task formulation, the game evaluation captures performance
variation that is also reflected in real-world visual reasoning.

\textbf{Thinking improves performance across model families.}
We compare thinking and non-thinking configurations for four models from different
families, covering both open-weight and proprietary systems. As shown in
Figure~\ref{fig:thinking_analysis}, all four models improve on the full 1,400-sample
benchmark. Qwen3.5-2B, Step3-VL-10B, and Gemma4-31B-it gain $6.2$, $6.0$, and
$11.1$ percentage points, respectively, while GPT-5.6-Sol gains $41.5$ points,
increasing from $28.3\%$ to $69.8\%$. The larger gains of Gemma4-31B-it and
GPT-5.6-Sol suggest that additional inference-time computation may be more
useful for models with stronger visual reasoning capabilities. Such models
may better use the extra reasoning budget to explore alternatives, verify
intermediate states, and revise earlier decisions. The smaller gains of
the other two models may partly reflect errors in visual understanding or
the initial problem representation that additional reasoning steps alone
cannot always correct.

\begin{figure}[t]
    \centering
    \includegraphics[width=\linewidth]{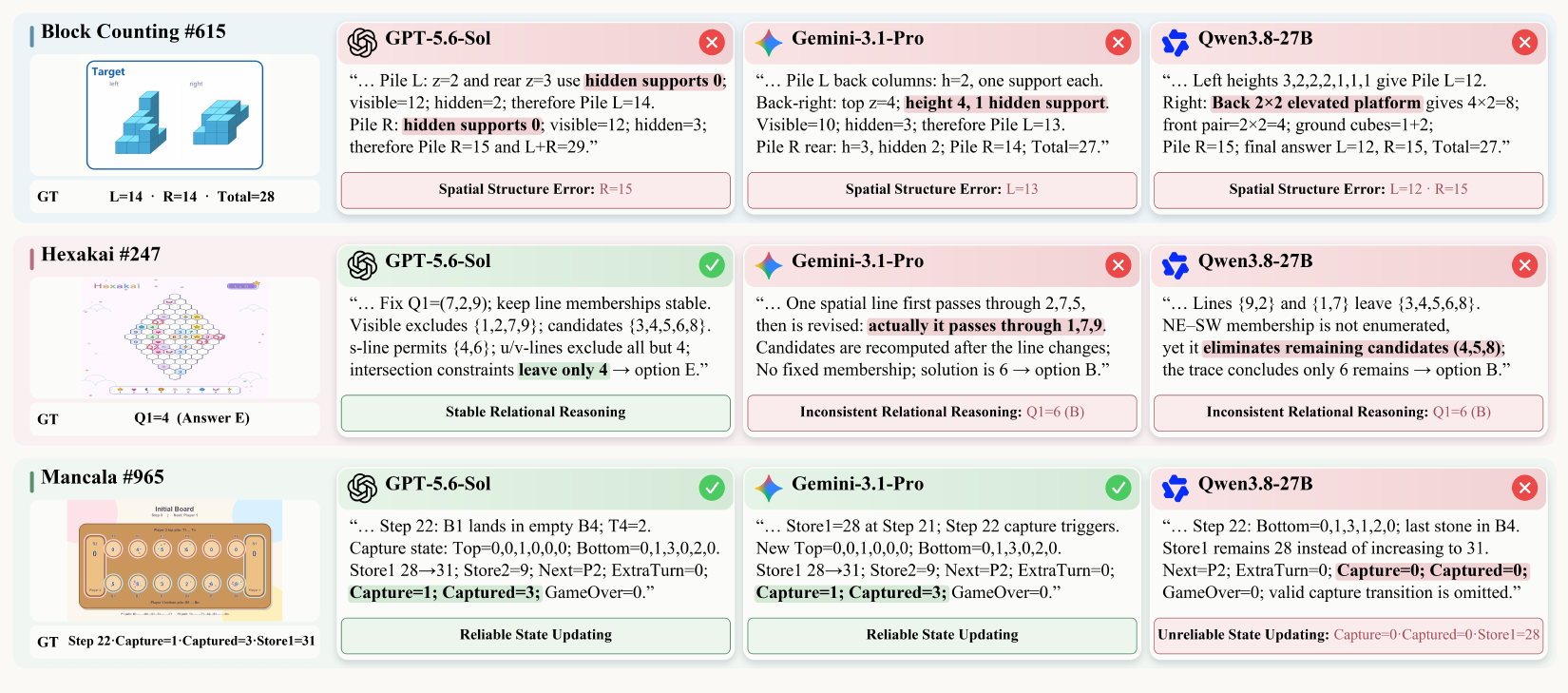}
    \vspace{-5mm}
    \caption{\textbf{Failure cases on Game2Real-Bench.}
    Model reasoning traces are compared with verified solutions for Block
    Counting, Hexakai, and Mancala. These cases illustrate three recurring
    failures: misinterpreting spatial structure, inconsistent relational
    reasoning, and unreliable state updating.}
    \label{fig:case_study}
\end{figure}

\textbf{Failure Case Analysis.}
We examine failure cases across all tasks and summarize three recurring error
patterns, illustrated by representative examples from Block Counting, Hexakai,
and Mancala in Figure~\ref{fig:case_study}.
\textbf{\textit{Misinterpreting Spatial Structure.}}
Models may form an incorrect spatial representation before reasoning begins.
In Block Counting, they fail to recover hidden cubes and column structures from
the 2D projection, leading to an incorrect 3D configuration before counting.
\textbf{\textit{Inconsistent Relational Reasoning.}}
Models may identify the relevant structure initially but fail to preserve its
relations during deduction. In Hexakai, Gemini-3.1-Pro changes constraint-line
memberships, while Qwen3.8-27B eliminates candidates without verifying the
relations.
\textbf{\textit{Unreliable State Updating.}}
Models may follow a sequence correctly for several steps but fail to apply a
required state transition. In Mancala, Qwen3.8-27B tracks the board state
through most of the trajectory but misses the capture update when its condition
is satisfied. These cases illustrate how errors can arise at different stages
of visual reasoning, from spatial representation to relation maintenance and
state transition.

\section{Limitations}

Game2Real-Bench focuses on four capability groups and connects toy-game tasks
with corresponding real-world visual reasoning settings. Several directions
remain for future work.
\textit{Broader task coverage.}
The current benchmark includes 18 toy-game tasks and 400 real-world samples.
Although these tasks cover diverse reasoning requirements, they do not span the
full range of visual reasoning in natural environments. Future extensions can
include additional task types and richer visual settings while preserving the
game--real alignment principle.
\textit{Broader transfer settings.}
Our transfer study focuses on one-shot in-context learning.
Game2Real-Dataset provides nearly 24K verified game samples, which can support
future studies of multi-shot prompting, supervised fine-tuning, and
reinforcement learning for game-to-real transfer.
\textit{Behavioral correspondence.}
The correlation between game and real-world performance reflects consistency
at the model level, but does not establish that the two settings rely on the
same internal reasoning processes. Further analysis can investigate how shared
reasoning requirements are represented and used across the two settings.

\section{Conclusion}

We introduce \textbf{Game2Real-Bench}, a benchmark that pairs controlled
toy-game tasks with real-world visual reasoning tasks to examine whether
abilities revealed in toy games are reflected in real-world performance and can
transfer across settings. Across 25 VLMs, toy-game and real-world performance is
strongly correlated, while toy-game tasks are generally more challenging and
provide clearer separation among current models. Capability-level results
further show that shared reasoning requirements can exhibit different
difficulty profiles when instantiated in the two settings. One-shot
game-to-real in-context learning also reveals selective transfer, with constraint
reasoning benefiting for several models while fine-grained perception
consistently declines. We also study the effects of thinking mode and analyze
representative failure cases. These results show that toy games provide an
informative setting for probing real-world visual reasoning and cross-setting
transfer. Together, \textbf{Game2Real-Bench} and
\textbf{Game2Real-Dataset} provide a verifiable evaluation framework and a
scalable data resource for advancing the study of game-to-real transfer in
visual reasoning.

\clearpage

\bibliography{iclr2027_conference}
\bibliographystyle{iclr2027_conference}

\appendix


\clearpage
\begingroup
\raggedbottom

\captionsetup[table]{
    font=normalsize,          
    labelfont=bf,             
    format=plain,            
    justification=justified, 
    singlelinecheck=false,   
    margin=0pt,              
    skip=5pt                 
}

\section{Detailed Benchmark Specification}
\label{g2r:app:specification}

We describe the game tasks in Game2Real-Bench and the accompanying
Game2Real-Dataset, including their taxonomy, sample counts, representative
examples, and QA-family coverage.


\subsection{Task Taxonomy and QA Formulation}
\label{g2r:app:taxonomy}

The 18 toy-game tasks are organized into four capability groups:
\textit{Fine-grained Perception} (C1), \textit{Spatial Reasoning} (C2),
\textit{Constraint Reasoning} (C3), and \textit{Sequential Planning} (C4).
Each task is assigned to the group that best reflects its primary
reasoning demand. Solving a task may also involve other capabilities,
such as perception during planning or constraint satisfaction in
spatial reasoning.

A \textit{task} defines a game environment with explicit rules and
solution criteria, while a \textit{QA family} specifies the reasoning
target and output format within that task. For example, Rolling Cube
includes questions about the top face, selected faces, or all six
faces after the roll sequence, as well as face states at intermediate
checkpoints. Each \textit{sample} instantiates a QA family with one or
more visual observations, a question, and a verified answer. Its
instructions specify the relevant notation, coordinate system,
and output conventions.

Table~\ref{g2r:tab:task_inventory} summarizes the task coverage and
sample counts for Game2Real-Dataset and the toy-game evaluation set
of Game2Real-Bench. Game2Real-Dataset contains 23,893 verified samples
across 65 QA families, while the independently generated evaluation
set contains 1,000 samples: 600 open-ended and 400 single-choice
questions. Both cover all 18 tasks. The \textit{Images} column counts
all visual inputs for each evaluation sample, including reference
views, auxiliary observations, and separately provided answer-option
images where applicable.

\begin{table}[H]
\centering
\caption{\textbf{Task inventory and sample counts.}
Dataset columns report QA families and samples in Game2Real-Dataset, while
evaluation columns report open-ended and single-choice samples in the toy-game
set. Images reports the number of visual inputs per evaluation sample, including
separately provided answer candidates.}
\label{g2r:tab:task_inventory}
\vspace{0.4em}

\small
\setlength{\tabcolsep}{5.5pt}
\renewcommand{\arraystretch}{1.08}

\begin{tabularx}{\textwidth}{
@{}
>{\raggedright\arraybackslash}m{0.30\textwidth}
Y Y Y Y Y
@{}
}
\toprule

& \multicolumn{2}{c}{\textbf{Dataset}}
& \multicolumn{2}{c}{\textbf{Evaluation}}
& \\

\cmidrule(lr){2-3}
\cmidrule(lr){4-5}

\textbf{Task}
& \textbf{Families}
& \textbf{Samples}
& \textbf{Open}
& \textbf{Choice}
& \textbf{Images}
\\

\midrule

\rowcolor{gray!10}
\multicolumn{6}{l}{\textbf{C1: Fine-grained Perception}} \\

Missing Tiles
& 3 & 518 & 45 & 18 & 1, 7 \\

Hidden Object
& 4 & 720 & 47 & 29 & 1, 2 \\

Dot Shape
& 3 & 480 & 0 & 24 & 7 \\

Color Swap
& 4 & 600 & 0 & 15 & 7 \\

\addlinespace[2pt]

\rowcolor{gray!10}
\multicolumn{6}{l}{\textbf{C2: Spatial Reasoning}} \\

Cube Net
& 3 & 1,200 & 45 & 19 & 1 \\

Cylinder Coloring
& 1 & 720 & 0 & 24 & 8 \\

Block Counting
& 4 & 1,079 & 29 & 29 & 1 \\

Rolling Cube
& 4 & 1,796 & 60 & 16 & 2 \\

\addlinespace[2pt]

\rowcolor{gray!10}
\multicolumn{6}{l}{\textbf{C3: Constraint Reasoning}} \\

Hexakai
& 6 & 1,680 & 37 & 27 & 1 \\

Tapa
& 4 & 2,140 & 45 & 30 & 1 \\

Battleship
& 4 & 1,440 & 29 & 15 & 1 \\

Block Island
& 4 & 1,560 & 35 & 32 & 1 \\

Pipe
& 3 & 1,560 & 40 & 15 & 1 \\

\addlinespace[2pt]

\rowcolor{gray!10}
\multicolumn{6}{l}{\textbf{C4: Sequential Planning}} \\

Laser
& 4 & 1,320 & 30 & 17 & 1 \\

KeyPath Quest
& 3 & 2,160 & 39 & 25 & 1 \\

Growing Trail
& 3 & 1,920 & 46 & 18 & 1 \\

Mancala
& 4 & 1,680 & 40 & 15 & 1 \\

Rush Exit
& 4 & 1,320 & 33 & 32 & 1 \\

\midrule

\textbf{Total}
& \textbf{65}
& \textbf{23,893}
& \textbf{600}
& \textbf{400}
& -- \\

\bottomrule
\end{tabularx}
\end{table}


\subsection{Task Definitions and Representative Examples}
\label{g2r:app:task_definitions}

Each task is accompanied by a representative QA from the toy-game
set. Questions are shortened for readability, and reference answers
are retained. Visual inputs are omitted here; sample IDs identify
records containing the complete images and prompts.

\subsubsection{Fine-grained Perception}

\gTwoRCompactCase{Missing Tiles}{402}
{Complete a missing tile by matching its connectors to neighboring ports,
with no extra connections. Port letters follow U, R, D, L order.}
{Which ports must the tile placed in the yellow gap R2C6 have?}
{R2C6=UL}

\gTwoRCompactCase{Hidden Object}{446}
{Count the requested animal classes, including recognizable partially
occluded instances. Region-restricted queries include only icons entirely
inside the region.}
{How many sheep and cows are present in the full scene?}
{Sheep=8; Cows=5}

\gTwoRCompactCase{Dot Shape}{151}
{Match the closed contour formed by the highlighted points, preserving
cyclic order and relative geometry. Translation and uniform scaling
are allowed; other transformations follow the question.}
{Which option A--F preserves the point identities, cyclic order,
and closed edge correspondence?}
{B}

\gTwoRCompactCase{Color Swap}{228}
{Apply the color mapping simultaneously to the selected pieces,
using their original colors and leaving other pieces unchanged.}
{Swap red with green on every piece; leave other colors unchanged.
Which option A--F shows the correct final model?}
{B}

\subsubsection{Spatial Reasoning}

\gTwoRCompactCase{Cube Net}{508}
{Fold the cube net while preserving face adjacency and sticker
orientation. Read sticker coordinates in the reference frame
specified by the question.}
{On the face opposite the blank 3D face, report R1C1, R2C2,
and R3C3 as drawn in the unfolded net.}
{R1C1=Green; R2C2=Red; R3C3=Green}

\gTwoRCompactCase{Cylinder Coloring}{177}
{Infer a missing orthographic view by matching cylinder positions,
layers, colors, and diameters across projections. The schematic
views retain all cylinders.}
{Given the Top View and Right View, which candidate A--F is the
correct Front View?}
{A}

\gTwoRCompactCase{Block Counting}{598}
{Count the complete cube piles, including required hidden supports
but excluding additional fully invisible cubes.}
{Report the left (L), right (R), and total cube counts,
including hidden supports.}
{L=5; R=7; Total=12}

\gTwoRCompactCase{Rolling Cube}{543}
{Track each $90^{\circ}$ roll while preserving physical face identities
and sticker orientation. Front denotes the face pointing toward
board South.}
{Report the Top and Front center symbols after steps 3, 7, and 10
of the displayed roll sequence.}
{Step3Top=Circle; Step3Front=Star; Step7Top=Square; Step7Front=Plus;
Step10Top=Square; Step10Front=Star}

\subsubsection{Constraint Reasoning}

\gTwoRCompactCase{Hexakai}{627}
{Complete the hexagonal grid using the displayed token-to-digit
legend, with no repeated digit along a straight line in any of
the three directions.}
{Solve the unique puzzle and report the digits at Q1, Q2, Q3, and Q4.}
{Q1=8; Q2=2; Q3=4; Q4=1}

\gTwoRCompactCase{Tapa}{666}
{Satisfy the neighboring shaded-run clues while keeping clue cells
white, shaded cells orthogonally connected, and every $2\times2$
block from being fully shaded.}
{Solve the unique puzzle and report black or white for R2C7, R4C6,
R6C3, R8C2, R6C5, R5C8, and R3C8.}
{R2C7=white; R4C6=white; R6C3=black; R8C2=black; R6C5=black;
R5C8=white; R3C8=white}

\gTwoRCompactCase{Battleship}{709}
{Place straight ships to satisfy the fleet inventory, row and
column totals, revealed clues, and contact restrictions.}
{List all cells of the unique topmost ship of length 3
in reading order.}
{ShipCoordinates=R4C5,R5C5,R6C5}

\gTwoRCompactCase{Block Island}{739}
{Place one L, I, T, or S tetromino per region. Shaded cells must be
edge-connected, with no filled $2\times2$ block; edge-touching
tetrominoes across regions must have different types.}
{Solve the unique board and list all shaded cells in Region01
and Region02 in row-major order.}
{Region01=R4C3,R4C4,R4C5,R4C6; Region02=R2C6,R2C7,R3C7,R4C7}

\gTwoRCompactCase{Pipe}{773}
{Rotate only dry pipes. Water propagates and newly wet pipes lock
after each operation; initially connected pipes are already locked.
Each operation applies one, two, or three clockwise quarter-turns.}
{Return the complete minimum plan of three ordered operations
that reaches the outlet, using RrCc=xk notation.}
{R6C5=x1; R4C4=x3; R3C2=x3}

\subsubsection{Sequential Planning}

\gTwoRCompactCase{Laser}{813}
{Orient mirrors so the ray visits relays alphabetically and reaches
the gem from the required direction. Every submitted mirror
must be used.}
{Return a complete optimal mirror configuration for R5C3, R6C2,
R6C3, R7C2, and R7C7.}
{R5C3=/; R6C2=/; R6C3=/; R7C2=\textbackslash{}; R7C7=/}

\gTwoRCompactCase{KeyPath Quest}{861}
{Find a route respecting walls and key-before-door dependencies.
Collected keys are retained after crossing their matching doors.}
{Return a globally shortest 46-move U/D/L/R route from S to E,
collecting every required key and crossing each matching door legally.}
{R D R R D L D L D D L U U U R L D D D D D R U R R R
U L L U R U R U R D D R U R D D L D D R}

\gTwoRCompactCase{Growing Trail}{907}
{Track the moving body: ordinary moves release the tail, while
eating fruit retains it. Avoid walls and body cells other than
a tail released on that step.}
{Collect all fruits in any legal order and finish at the yellow
goal. Return a globally shortest 10-move U/D/L/R route.}
{U U L D L L U U R D}

\gTwoRCompactCase{Mancala}{928}
{Execute moves under the sowing, capture, extra-turn, and
terminal-sweep rules. Termination takes precedence over extra turns;
event flags describe the last executed move.}
{Execute B5 followed by B6. Return the complete board state
and event flags after move 2.}
{Top=7,3,0,1,8,0; Bottom=3,0,0,7,0,0; Store1=2; Store2=0;
Next=P1; Step=2; ExtraTurn=1; Capture=0; Captured=0; GameOver=0}

\gTwoRCompactCase{Rush Exit}{994}
{Slide vehicles along their fixed axes without overlap. Each command
costs one move regardless of distance; only the target can exit,
ending execution at its first boundary crossing.}
{Move target C1 through the left EXIT using a globally shortest
plan of five semicolon-separated commands.}
{C5 LEFT1; C3 UP3; C5 RIGHT1; C7 UP1; C1 LEFT5}


\clearpage
\subsection{QA-family Overview}
\label{g2r:app:qa_families}

The 65 QA families vary the query scope and requested output within
each task. They cover local attributes, aggregate counts, complete
configurations, intermediate states, and action sequences.
Table~\ref{g2r:tab:families} summarizes this coverage by task.
Some families support both open-ended and single-choice variants,
with the exact answer format specified by the sample instructions.

\begin{table}[H]
\centering
\small
\setlength{\tabcolsep}{5pt}
\renewcommand{\arraystretch}{1.15}

\caption{\textbf{QA-family coverage by task.}
Each row summarizes the query targets represented in Game2Real-Dataset.
The number of families per task is reported in
Table~\ref{g2r:tab:task_inventory}.}
\label{g2r:tab:families}

\begin{tabular}{
    @{}
    >{\RaggedRight\arraybackslash}p{0.245\linewidth}
    >{\RaggedRight\arraybackslash}p{0.70\linewidth}
    @{}
}
\toprule
\textbf{Task} & \textbf{QA-family coverage} \\
\midrule

\rowcolor{black!5}
\multicolumn{2}{l}{\strut\textbf{C1: Fine-grained Perception}} \\
Missing Tiles
& Complete connector patterns; side states; port lists and degrees. \\
Hidden Object
& Whole-scene and region-specific counts across target classes. \\
Dot Shape
& Contour matching; vertex counts; counts by contour layer. \\
Color Swap
& Transformed models; full or partial color histograms;
changed and unchanged piece counts. \\

\addlinespace[3pt]
\rowcolor{black!5}
\multicolumn{2}{l}{\strut\textbf{C2: Spatial Reasoning}} \\
Cube Net
& Face identification; complete sticker patterns;
selected sticker cells. \\
Cylinder Coloring
& Missing-view selection from orthographic projections. \\
Block Counting
& Pile totals; occupied columns; hidden supports; height histograms. \\
Rolling Cube
& Final top, selected, or all six faces; intermediate checkpoints. \\

\addlinespace[3pt]
\rowcolor{black!5}
\multicolumn{2}{l}{\strut\textbf{C3: Constraint Reasoning}} \\
Hexakai
& Joint values of one to six queried cells. \\
Tapa
& Selected cell states or the complete shaded-cell set. \\
Battleship
& Selected cell states or the complete fleet configuration. \\
Block Island
& Tetromino placements in selected regions or the complete board. \\
Pipe
& Minimum ordered rotation plans; operation counts; wet-cell progression. \\

\addlinespace[3pt]
\rowcolor{black!5}
\multicolumn{2}{l}{\strut\textbf{C4: Sequential Planning}} \\
Laser
& Selected or complete mirror assignments; candidate selection;
trajectory key points. \\
KeyPath Quest
& Shortest routes; key-collection order; timed key and door events. \\
Growing Trail
& Shortest routes; fruit-collection order and collection steps. \\
Mancala
& Final or checkpoint states; event counts and timelines. \\
Rush Exit
& Minimum-command plans; command statistics;
moved or repeatedly moved vehicles. \\

\bottomrule
\end{tabular}
\end{table}

\clearpage
\endgroup


\clearpage
\begingroup
\raggedbottom
\captionsetup[table]{
    font=normalsize,
    labelfont=bf,
    format=plain,
    justification=justified,
    singlelinecheck=false,
    margin=0pt,
    skip=5pt
}
\providecolor{gTwoRPerception}{HTML}{92500B}
\providecolor{gTwoRSpatial}{HTML}{176263}
\providecolor{gTwoRConstraint}{HTML}{465681}
\providecolor{gTwoRPlanning}{HTML}{75072D}

\section{Benchmark Construction Details}
\label{g2r:app:construction}

We detail the construction pipeline through Rolling Cube, focusing
on how the generator and verifier track physical faces and their
sticker patterns. We then use an IKEA assembly example to show how
aligned instructions and video frames become a multi-image question
with a reference answer.

\subsection{Procedural Generation}
\label{g2r:app:generation}

\paragraph{Initial face patterns.}
The generator samples a $3\times3$ sticker grid for each of the six
faces. Each grid contains all six symbols in the selected vocabulary,
with three symbols appearing twice. During sampling, grids are compared
under rotations and reflections to avoid equivalent patterns on
separate faces. The evaluation uses geometric symbols; dataset
construction also supports digits and letters. We retain the initial layouts and
their vocabulary separately from the answers derived after rolling.

\paragraph{Movement sequence and visual inputs.}
The generator samples a start cell and extends a path through
orthogonally adjacent cells. Configuration profiles control the
board size, number of rolls, and changes between horizontal and
vertical movement. Proposed paths must remain within the board and
satisfy the profile's turn and revisit conditions. Each movement
represents a $90^{\circ}$ roll, not a translation or an in-place
rotation. The instance is rendered into two inputs: a numbered path
on the board and a panel containing the six initial face layouts.
Their roles and order are stated in the prompt, with the path and
face grids preserved in the verification metadata.

\paragraph{Queries and distractors.}
QA construction selects the requested checkpoint and world-facing
side. Full-grid questions in Game2Real-Dataset require the entire
oriented pattern; other variants query selected faces, the complete
six-face state, or intermediate checkpoints. For a six-option grid
question, the distractor pool includes rotations and reflections of
the target grid, alternative face states, and local symbol changes.
Duplicate grids are removed before options are selected and shuffled.
These alternatives distinguish an incorrect face from a correct
face with the wrong orientation.

\subsection{Task-Specific Verification}
\label{g2r:app:verification}

\paragraph{Replaying the cube orientation.}
The checker reads the initial face grids and ordered moves from the
structured state and recomputes the result without consulting the
stored answer label. Let $x$, $y$, and $z$ point East, North, and
upward, respectively. The implementation represents cube orientation
by an integer rotation matrix $R_k$. For each move $d_k$ on the
displayed path, we update $R_k$ as
\begin{equation}
    R_0=I, \qquad R_k=Q(d_k)R_{k-1}.
    \label{g2r:eq:rolling_update}
\end{equation}
For East and North rolls, the matrices are
\begin{equation}
    Q(E)=
    \begin{pmatrix}
       0&0&1\\ 0&1&0\\ -1&0&0
    \end{pmatrix},
    \qquad
    Q(N)=
    \begin{pmatrix}
       1&0&0\\ 0&0&1\\ 0&-1&0
    \end{pmatrix},
    \label{g2r:eq:rolling_matrices}
\end{equation}
with $Q(W)=Q(E)^{\mathsf T}$ and $Q(S)=Q(N)^{\mathsf T}$.
The checker applies the resulting orientation to the reconstructed
sticker state and reads the requested grid using the defined
world-face axes. For Top, rows run from board North to South and
columns from West to East. Checking a full grid compares all nine
positions after the full roll sequence, rather than only the
symbol at the face center.

\paragraph{A full-pattern example.}
We use dataset sample \texttt{GBT-COT-004927}, which asks for the
final Top grid after the seven rolls $E,N,E,S,E,N,E$. Replaying
this sequence gives
\begin{equation}
    R_7=
    \begin{pmatrix}
       0&0&-1\\ 0&1&0\\ 1&0&0
    \end{pmatrix},
    \qquad
    G_{\mathrm{Top}}=
    \begin{pmatrix}
       6&6&4\\ 1&3&2\\ 4&2&5
    \end{pmatrix}.
    \label{g2r:eq:rolling_example}
\end{equation}
The initial East face becomes Top, but its center symbol, 3,
alone cannot identify the answer. Options A, B, D, E, and F all
share that center and the same symbol counts. Only A matches the
recomputed arrangement. Table~\ref{g2r:tab:rolling_verification}
shows how the checker distinguishes these actual candidate grids.

\begin{table}[H]
\centering
\caption{\textbf{Candidate verification for Rolling Cube.}
The options are taken from Game2Real-Dataset sample
\texttt{GBT-COT-004927}; slashes separate grid rows. Rotations are
clockwise relative to the recomputed Top grid. Only A agrees at
every position under the specified reading convention.}
\label{g2r:tab:rolling_verification}
\small
\setlength{\tabcolsep}{6pt}
\renewcommand{\arraystretch}{1.14}
\begin{tabular}{@{}lllr@{}}
\toprule
\textbf{Option} & \textbf{Candidate grid} & \textbf{Relation to target} & \textbf{Valid} \\
\midrule
\rowcolor{gTwoRSpatial!8}
A & \texttt{6 6 4 / 1 3 2 / 4 2 5} & Exact arrangement & Yes \\
B & \texttt{5 2 4 / 2 3 1 / 4 6 6} & $180^{\circ}$ rotation & No \\
C & \texttt{4 5 4 / 6 1 1 / 2 6 3} & Final Bottom grid & No \\
D & \texttt{4 2 5 / 6 3 2 / 6 1 4} & $270^{\circ}$ rotation & No \\
E & \texttt{4 1 6 / 2 3 6 / 5 2 4} & $90^{\circ}$ rotation & No \\
F & \texttt{4 6 6 / 2 3 1 / 5 2 4} & Left--right reflection & No \\
\bottomrule
\end{tabular}
\end{table}

\paragraph{Answer recomputation from structured state.}
The validation interface removes declared answer labels and correctness
flags before invoking the task checker. Rolling Cube reconstructs its result
from \texttt{initial\_faces} and \texttt{moves}, rather than reading a saved
final layout or candidate-validity flag. Each option is parsed into a
$3\times3$ grid and compared with the recomputed Top grid. Acceptance requires
exactly one valid option and agreement with the exported answer. The checker
reuses the generator's rotation and layout functions, verifying label-independent
state recomputation under the shared simulation implementation.

\paragraph{Auditing the exported question.}
Release checks address the interface between the state and the
question: the two input images must have the correct roles, path
markers must be distinguished from direction names, and face names
must denote world-facing positions at the requested checkpoint.
The stored layout, candidate order, and exported label must agree.
For open-ended variants, the reference is also passed through the
answer parser and evaluator to check field names and symbol
normalization. The parser round trip checks output formatting, while state replay checks
consistency with the implemented task dynamics.

\paragraph{Other tasks and collection-level checks.}
Other tasks use the same separation between generation, answer
checking, and export auditing. Object annotations support counting,
constraint solvers check complete assignments, and dynamic tasks
replay actions with their state-dependent effects. Uniqueness or
optimality is checked when the task requires it; Rolling Cube
instead specifies a fixed sequence with a determined outcome.
Collection-level checks remove benchmark--dataset overlaps using
semantic fingerprints and image hashes.

\subsection{Real-World Curation and Game-to-Real Alignment}
\label{g2r:app:realworld}

\paragraph{Scope of game--real alignment.}
Game2Real-Bench connects procedural games with real-world questions
through shared reasoning requirements. Alignment guides curation and
capability-level reporting rather than requiring a one-to-one correspondence
between tasks or samples. The two settings retain different visual
evidence, rules, and answer spaces. For example, Cube Net and the IKEA
assembly questions both relate a two-dimensional specification to a
three-dimensional configuration. Their connection concerns spatial
interpretation, while cube folding and assembly-state recognition retain
distinct operations, input formats, and solution criteria.

\paragraph{Primary annotations and task coverage.}
Each real-world sample has a primary toy-task annotation recording its
designated game--real connection. This annotation is not an exhaustive list
of the operations needed to solve the question. Likewise, capability labels
identify primary demands rather than mutually exclusive skills: Rolling Cube
is grouped under spatial reasoning but also requires sequential state
tracking. The primary annotations cover 14 game task types, whereas the
full toy-game evaluation retains all 18 tasks. The other four tasks have no
retained real-world sample with that primary annotation. Accordingly, the
full-suite correlation assesses model-level performance correspondence;
it is not an estimate of the predictive value of each individual task link.

\paragraph{Source-specific construction.}
Table~\ref{g2r:tab:real_sources} lists the real-world sources.
We retain 246 original QA samples and lightly reformulate 72 while
preserving their answers. New annotation-derived questions comprise
27 SPair-71k keypoint-matching samples and five IKEA assembly-state
samples. Another 50 SpatiaLQA samples retain the official tasks with
restored action-content and precondition graphs. The IKEA example illustrates the complete construction process.

\begin{table}[!htbp]
\centering
\caption{\textbf{Real-world sources and construction.}
Counts refer to retained Game2Real-Bench samples. Original QA is kept
where suitable; light reformulation preserves source answers, while
annotation-derived QA introduces questions grounded in source annotations.}
\label{g2r:tab:real_sources}
\small
\setlength{\tabcolsep}{5pt}
\renewcommand{\arraystretch}{1.15}
\begin{tabular}{@{}
>{\RaggedRight\arraybackslash}p{0.405\linewidth}
r
>{\RaggedRight\arraybackslash}p{0.50\linewidth}@{}}
\toprule
\textbf{Source} & \textbf{$N$} & \textbf{Construction} \\
\midrule
GQA~\citep{hudson2019gqa} & 12
& 10 original QA; 2 lightly reformulated. \\
OmniSpatial~\citep{jia2026omnispatial} & 61
& Original visual and spatial reasoning QA. \\
SPair-71k~\citep{min2019spair} & 27
& QA derived from keypoint correspondences. \\
Visual-Jigsaw~\citep{wu2026visual} & 40
& Lightly reformulated completion questions. \\
CAPTURe-real~\citep{pothiraj2025capture} & 30
& Lightly reformulated counting questions. \\
All-Angles-Bench~\citep{yeh2026seeing} & 20
& Original synchronized-view questions. \\
IKEA Manuals at Work~\citep{liu2024ikea} & 5
& QA derived from aligned diagrams and video. \\
SpatiaLQA~\citep{xie2026spatialqa} & 50
& Official tasks with restored dependency graphs. \\
SSI-Bench~\citep{yang2026thinking} & 50
& Original geometric and topological ranking QA. \\
VisualTrans~\citep{ji2025visualtrans} & 91
& Original transformation questions. \\
ReasonMap~\citep{feng2026reasonmap} & 14
& Original transit-route questions. \\
\midrule
\textbf{Total} & \textbf{400} & \textbf{100 samples per capability.} \\
\bottomrule
\end{tabular}
\end{table}

\paragraph{Locating the assembly step.}
For IKEA Manuals at Work~\citep{liu2024ikea}, official alignments identify
the instruction diagram, source video, and assembly step. We inspect the
aligned material for a visible change in part arrangement, connection,
or orientation and select distinguishable pre-step and post-step states.
The five retained items cover \texttt{applaro}, \texttt{stig},
\texttt{vesken}, \texttt{laiva}, and \texttt{satsumas}.
Each question uses one instruction and one video, keeping object identity
and background consistent across pre-step distractors and post-step candidates.

\paragraph{Temporal selection and frame extraction.}
Each item contains four pre-step and four post-step frames,
spaced one second apart within each group. For \texttt{applaro}
step~00, the two groups start at 191.60 and 256.48 seconds.
Table~\ref{g2r:tab:ikea_timestamps} records all eight timestamps
and shuffled labels. We sample distinguishable states instead
of sampling uniformly over the full action, preserving the
exact frame selection for reconstruction from the source video
and its aligned assembly step.

\begin{table}[H]
\centering
\caption{\textbf{Frame selection for the IKEA \texttt{applaro} example.}
Timestamps are measured in seconds from the beginning of the source video.
Each group contains four frames; labels identify their shuffled positions
after the instruction diagram in the final nine-image input.}
\label{g2r:tab:ikea_timestamps}
\small
\setlength{\tabcolsep}{9pt}
\renewcommand{\arraystretch}{1.15}
\begin{tabular}{@{}rc@{\hspace{3em}}rc@{}}
\toprule
\rowcolor{gTwoRSpatial!8}
\multicolumn{2}{c}{\textbf{Pre-step frames}}
& \multicolumn{2}{c}{\textbf{Post-step frames}} \\
\cmidrule(lr){1-2}\cmidrule(lr){3-4}
\textbf{Time (s)} & \textbf{Label} & \textbf{Time (s)} & \textbf{Label} \\
\midrule
191.60 & F2 & 256.48 & F8 \\
192.60 & F7 & 257.48 & F1 \\
193.60 & F5 & 258.48 & F3 \\
194.60 & F6 & 259.48 & F4 \\
\bottomrule
\end{tabular}
\end{table}

\paragraph{Spatial cropping and input assembly.}
After decoding the selected frames, we remove the timestamp-bearing
lower image band to avoid revealing their temporal order. The eight
frames are shuffled and labeled F1--F8, with the instruction
placed first. Each question therefore contains nine images: one diagram
and eight candidates. Original timestamps, state groups, display
positions, and the source-video hash remain in the metadata but are
withheld from the model. Image-role instructions identify the diagram
and the candidate corresponding to each displayed label.

\paragraph{Question and answer annotation.}
The question asks for the four frames showing the closest valid post-step
3D state. For \texttt{applaro}, the stored answer is
\texttt{["F8","F1","F3","F4"]}, evaluated as an unordered set.
Pre-step frames provide distractors from the same scene. Review checks
agreement between the diagram and post-step part configurations, and
whether the two state groups remain distinguishable after cropping.
The partial alignment with Cube Net captures shared 2D-to-3D part
correspondence while retaining the assembly task's distinct operations.

\paragraph{Other annotation-derived questions and review.}
SPair-71k questions place a red marker at an annotated source keypoint
and four labeled markers on the target image. The answer follows from
the semantic correspondence. We retain nine samples from each of the
articulation, high-scale-change, and high-viewpoint-change subsets.
Across sources, automated checks validate fields, image references,
labels, and source annotations. All 400 selected samples undergo manual
review of their visual evidence, reference answers, and game--real
mappings, with source-specific annotations providing the basis for
checking each reference answer against its visual inputs.

\clearpage
\endgroup


\section{Evaluation Protocol}
\label{g2r:app:evaluation}

We detail the inference configurations, model inputs, and scoring procedures
for both evaluation settings. Shared prompt construction preserves
task-specific output conventions, while reference answers and evaluator
metadata remain separate from the target input supplied to each model.

\subsection{Model and Inference Configurations}
\label{g2r:app:model_configurations}

\textbf{Model interfaces.}
Open-weight models are evaluated through VLMEvalKit adapters, with checkpoint,
chat-template, and visual preprocessing settings determined by the corresponding
run configuration~\citep{duan2024vlmevalkit}. Proprietary models are accessed
through API interfaces. Each model receives the source images and shared task
prompt, without external tools, reference solutions, or task-specific solvers.
Model-specific inference controls are summarized in
Appendix~\ref{g2r:app:thinking_configuration}.

\textbf{Thinking-mode generation for Qwen.}
For Qwen3.5 and Qwen3.8 with thinking enabled, we use a two-stage generation
budget of $32\mathrm{k}+512$ tokens. The initial stage allows up to
$32\mathrm{k}$ generated tokens. If this budget is exhausted without a final
answer, we continue the same attempt with an answer-format instruction and
up to 512 additional tokens, without reference answers or evaluator feedback.
Responses completed within the initial budget are evaluated directly. Both
paths use the same answer normalization and scoring procedures.

\textbf{Controlled in-context evaluation.}
The one-shot analysis evaluates Qwen3.5-2B, Qwen3.5-4B, Qwen3.5-9B, and
Qwen3.8-27B on the same 400 real-world targets. The controlled Qwen runner
uses vLLM with native multi-turn conversations and model-specific custom
prompts disabled (Table~\ref{g2r:tab:icl_configuration}). Paired zero-shot
and one-shot evaluations retain the same configuration for each model,
including the two-stage generation budget when thinking is enabled.
The paired runs differ only in demonstration context, not in inference mode.

\begin{table}[H]
\centering
\caption{\textbf{Configuration of the controlled Qwen in-context learning runner.}
Paired zero-shot and one-shot runs share these inference settings. In thinking
mode, the additional answer budget is used only when the initial budget is
exhausted without a final answer.}
\label{g2r:tab:icl_configuration}
\small
\setlength{\tabcolsep}{5pt}
\renewcommand{\arraystretch}{1.08}
\begin{tabularx}{\linewidth}{@{}l X l X@{}}
\toprule
\textbf{Setting} & \textbf{Value} & \textbf{Setting} & \textbf{Value} \\
\midrule
Backend & vLLM & Temperature & 0 \\
Sampling & Disabled & Context limit & 65,536 \\
Top-$p$ / top-$k$ & 0.95 / 20 & Presence penalty & 1.5 \\
Thinking budget & $32\mathrm{k}$ tokens & Answer budget & 512 tokens \\
Custom prompts & Disabled & Answer stage & Conditional \\
\bottomrule
\end{tabularx}
\end{table}

\subsection{Prompting and Input Formatting}
\label{g2r:app:prompting}

\textbf{Shared input construction.}
The \texttt{ImageMixedQA} and \texttt{RealWorldFinal} interfaces share one
prompt builder. Each input contains the stored image sequence followed by a
text block with optional task instructions from the \texttt{hint} field, the
question, available options, image-role instructions, and the answer-format
suffix. The target's reference answers, accepted alternatives, and evaluator
payloads are reserved for scoring and excluded from the prompt and its
image-role descriptions.

\textbf{Image roles and answer candidates.}
The prompt distinguishes question and reference images from images that
represent answer options. For example, Rolling Cube provides the board and
roll sequence separately from the initial cube layout, whereas Cylinder
Coloring supplies reference projections followed by candidate views. When
candidates are supplied as separate images, the prompt explicitly maps their
positions to option letters, preventing an image index from being confused
with an answer label. When the options are textual, all supplied views remain
question or reference images and are interpreted jointly in the stored order,
with their task-specific roles preserved.

\textbf{Answer-format instructions.}
The shared suffix requests a direct answer and ends with a question-type-specific
instruction, as shown in the prompt box. Open-ended questions retain their requested
schema, including field assignments, action sequences, rankings, and JSON
objects. For example, IKEA frame selections remain JSON arrays of frame
labels rather than single-choice option letters. Output format and model-side thinking mode are configured separately
at inference.

\begin{figure}[t]
\begin{tcolorbox}[
  colback=black!2, colframe=black!40, boxrule=0.4pt,
  arc=1mm, left=1.5mm, right=1.5mm, top=1mm, bottom=1mm,
  fontupper=\small, title={Shared answer instruction},
  colbacktitle=black!8, coltitle=black, fonttitle=\small\bfseries]
Answer directly. Do not explain your reasoning.\par
For single-choice questions, answer with one option letter only, such as A.\par
For multiple-answer questions, answer with all correct option letters
separated by commas, such as A,C,E.\par
For open questions, follow the required answer format exactly.\par
\textbf{Single-choice suffix:} This is a single-choice question.
Return one option letter only.\par
\textbf{Open-ended suffix:} This is an open question.
Return only the requested structured answer.
\end{tcolorbox}
\end{figure}

\textbf{One-shot conversation format.}
A game demonstration is represented as a user turn containing its images and
question, followed by an assistant turn containing only its verified final
answer. The real-world target is then supplied as a new user turn using the
same input construction as in zero-shot evaluation. Demonstrations therefore
provide solved examples without additional reasoning traces or target
annotations. The three assignments use seeds 11, 23, and 37, with identical
demonstration--target pairs across models. Only the demonstration context
changes; the target images, question, answer schema, and scoring procedure
remain the same across the paired evaluations.

\subsection{Toy-Game Answer Normalization and Scoring}
\label{g2r:app:toy_scoring}

\textbf{Schema-specific normalization.}
Toy-game predictions are evaluated against the answer schema attached to each
sample. For the 400 single-choice questions, the answer is mapped to one of
six valid option letters and compared with the reference label. For the 600
open-ended questions, deterministic parsers recover the requested values or
actions and pass them to the corresponding task evaluator. Normalization is
restricted to representations supported by that schema, such as whitespace,
letter case, recognized symbol aliases, and equivalent field separators;
it does not infer omitted values, repair an incorrect solution, or alter
the submitted sequence of actions.

\textbf{Structured answers.}
For field-based outputs, the parser checks the complete set of requested keys
and their associated values. Reordering independent field assignments is
permitted, while missing fields, duplicate keys, and unexpected fields are
rejected. Coordinate collections are normalized where their order is
irrelevant, whereas checkpoint order, vector entries, and action sequences
retain their prescribed meaning. For example, reversing the order of two
named count fields does not change an answer, but reversing two moves can
change the resulting state. The open-answer parser must consume the complete
submitted answer, so unsupported wrappers and surrounding explanations are
not silently ignored during schema validation to obtain a matching answer.

\textbf{Task-grounded correctness.}
Attribute, count, and queried-state answers require all requested components
to match their verified targets. Tasks admitting alternative solutions are
instead checked against their executable rules. For KeyPath Quest and
Growing Trail, the evaluator replays the submitted route, verifies legal
transitions and goal completion, and checks its length against the stored
optimum. Rush Exit similarly checks vehicle movements and the minimum number
of commands, rather than the total distance traveled. Laser verifies the
submitted mirror configuration through ray simulation, including the relay
order, target entry, and optimality conditions specified by the task.

\textbf{Order-sensitive state updates.}
Pipe predictions are evaluated as ordered rotation sequences. The evaluator
applies each operation to the current state, propagates water, and updates
which pipes are locked before considering the next operation. It therefore
does not reorder a submitted plan to make an otherwise invalid sequence
succeed. Mancala evaluation checks the requested board state and event fields
against the replayed transitions, including sowing, captures, extra turns,
and terminal sweeping. These checks require both the requested board components and the event
flags to agree with the replayed state; a partially correct configuration
does not receive credit.

\textbf{Sample scores.}
Each toy-game sample receives a binary score. A prediction is correct only
when it satisfies the answer schema and all required task conditions,
including optimality when requested. A valid alternative solution can
therefore receive full credit without matching the stored canonical answer
verbatim. Invalid or incomplete answers receive zero, and intermediate
agreement is not awarded partial credit. Parser and simulator diagnostics are retained separately for error analysis.

\subsection{Real-World Evaluation Protocols}
\label{g2r:app:real_scoring}

\textbf{Reference-independent answer extraction.}
Real-world responses are first converted to the output type specified by each
sample. The extractor removes recognized reasoning blocks and output
wrappers, prioritizes explicit final-answer markers, and recovers structured
values from JSON or Python-style lists when applicable. For choice and
numeric questions, type-specific fallbacks recover a candidate letter or
number when no explicit final-answer field is available. This stage uses the
response and question-side information, without access to reference answers
or accepted alternatives. The extracted prediction is then passed to the
appropriate evaluator in Table~\ref{g2r:tab:real_evaluation}.

\begin{table}[H]
\centering
\caption{\textbf{Scoring rules for the 400 real-world samples.}
Counts refer to the final selected set. All subsets use binary correctness
except SpatiaLQA, which contributes a normalized graph score.}
\label{g2r:tab:real_evaluation}
\small
\setlength{\tabcolsep}{5pt}
\renewcommand{\arraystretch}{1.12}
\begin{tabularx}{\linewidth}{@{}p{0.36\linewidth}r X@{}}
\toprule
\textbf{Subset or answer type} & \textbf{Samples} & \textbf{Scoring criterion} \\
\midrule
Single-choice subsets & 236 & Exact option-label match. \\
GQA short answers & 12 & Normalized short-answer match. \\
CAPTURe-real counts & 30 & Exact count correctness. \\
VisualTrans and IKEA frame selection & 8 & Exact selected labels, ignoring order. \\
SSI-Bench rankings & 50 & Exact match of the full ranking. \\
ReasonMap routes & 14 & Match to an accepted structured route. \\
SpatiaLQA dependency graphs & 50 & Mean of content and precondition F1. \\
\midrule
\textbf{Total} & \textbf{400} & \\
\bottomrule
\end{tabularx}
\end{table}

\textbf{Choice, short-answer, and counting tasks.}
Single-choice predictions are compared with the reference option label.
Short answers are compared after text normalization, and counting questions
require the predicted count to match the reference. When a sample contains
multiple accepted answers, matching any one of them is sufficient. For the
CAPTURe-real subset, this is an adaptation to exact count correctness rather
than the original sMAPE metric~\citep{pothiraj2025capture}. The count remains a discrete target, with no tolerance-based partial credit
for approximate answers.

\textbf{Frame selection and structural ranking.}
The reconstructed IKEA questions require selecting frames that match the
post-step state indicated by an assembly instruction. Their frame labels are
compared without regard to order, as are the corresponding multi-frame
selections from VisualTrans. The complete selection must match the reference;
missing or additional labels are incorrect. In contrast, SSI-Bench requires
an ordered permutation of structural elements. We use its taskwise
exact-match criterion, rather than pairwise ordering
accuracy~\citep{yang2026thinking}. Output normalization preserves the ranking,
including any tie-breaking convention stated in the original question.

\textbf{Transit-route answers.}
ReasonMap responses are parsed into ordered route sections containing the
line name, departure and arrival stops, and any requested intermediate-stop
counts or lists. The evaluator normalizes textual representations and
compares the complete route with the accepted reference routes. Section
order and the requested stop information are retained during comparison.
This produces a binary structured-route score for Game2Real-Bench, rather
than reproducing ReasonMap's full correctness and route-quality evaluation
or its graded map score~\citep{feng2026reasonmap}.

\textbf{SpatiaLQA step content and dependencies.}
We adopt SpatiaLQA's distinction between step-content correctness and
precondition correctness~\citep{xie2026spatialqa}. Each answer represents an
execution plan as a graph whose nodes contain action descriptions and whose
edges encode prerequisite steps. Predicted and reference steps are aligned
through a binary correspondence matrix and a maximum-cardinality one-to-one
matching. Content precision and recall count matched, extra, and missing
steps. After aligning step identities, precondition precision and recall
compare the predicted and reference dependency edges. Changing step identifiers or JSON key order does not change the plan,
provided that its action contents and dependencies are preserved.

The benchmark implementation uses rule-based action, object, and contextual
matching at a threshold of $0.72$, replacing the official GPT-4o-based
correspondence stage. Content and precondition F1 scores are then computed
from the resulting step alignment. For sample $i$, the benchmark uses
$s_i=(F_{c,i}+F_{p,i})/2$, where $F_{c,i}$ and $F_{p,i}$ denote content and
precondition F1, respectively. This averaging defines our benchmark-level
aggregation, rather than an additional official SpatiaLQA metric.
Predictions with no matched steps receive zero and remain in the
evaluation denominator.

\textbf{Aggregation across tasks and settings.}
For any evaluation set $\mathcal{S}$, we report
$\operatorname{Score}(\mathcal{S})=100\,|\mathcal{S}|^{-1}
\sum_{i\in\mathcal{S}}s_i$.
Each real-world capability contains 100 samples, so its overall score is
also the unweighted mean of the four capability scores. The joint benchmark
score instead averages over all 1,400 samples, assigning weights of
$1{,}000/1{,}400$ and $400/1{,}400$ to the toy-game and real-world sets.
All selected samples remain in the denominator. This quantity is accuracy
for binary-scored subsets and an average normalized score when the
SpatiaLQA graph scores are included.

\clearpage


\section{Details of Analysis Experiments}
\label{g2r:app:analysis_experiments}

We detail the data correspondence and score construction for cross-setting
correlation, demonstration selection and prompting for one-shot transfer,
and inference controls for thinking-mode evaluation. All analyses retain the
benchmark partitions and scoring procedures in Appendix~\ref{g2r:app:evaluation}.

\subsection{Cross-Setting Correlation}
\label{g2r:app:cross_setting_correlation}

\textbf{Correspondence between evaluation sets.}
The two evaluation sets share four capabilities but retain different sample
distributions. The toy-game set contains 178 fine-grained perception, 222 spatial
reasoning, 305 constraint reasoning, and 295 sequential planning samples;
each real-world capability contains 100 samples.
Table~\ref{g2r:tab:analysis_alignment} identifies their correspondence through
primary toy-task annotations. These labels describe shared reasoning
requirements, rather than paired renderings or one-to-one question matches.

\begin{table}[H]
\centering
\caption{\textbf{Task correspondence and demonstration-pool composition.}
Toy and Real report evaluation counts; source counts in parentheses sum to
Real in each row. Pool reports demonstration counts. Target-side primary task
labels define these associations, independently of ICL assignments; zero denotes
an absent primary alignment.}
\label{g2r:tab:analysis_alignment}
\small
\setlength{\tabcolsep}{4pt}
\renewcommand{\arraystretch}{1.08}
\begin{tabularx}{\linewidth}{@{}l r r >{\raggedright\arraybackslash}X r@{}}
\toprule
\textbf{Toy task} & \textbf{Toy} & \textbf{Real} & \textbf{Real-world sources} & \textbf{Pool} \\
\midrule
\rowcolor{black!5}\multicolumn{5}{l}{\textbf{C1: Fine-grained Perception}} \\
Missing Tiles & 63 & 40 & Visual-Jigsaw (40) & 12 \\
Hidden Object & 76 & 21 & OmniSpatial (21) & 12 \\
Dot Shape & 24 & 27 & SPair-71k (27) & 12 \\
Color Swap & 15 & 12 & GQA (12) & 12 \\
\addlinespace[2pt]
\rowcolor{black!5}\multicolumn{5}{l}{\textbf{C2: Spatial Reasoning}} \\
Cube Net & 64 & 5 & IKEA Manuals at Work (5) & 12 \\
Cylinder Coloring & 24 & 13 & All-Angles-Bench (13) & 12 \\
Block Counting & 58 & 30 & CAPTURe-real (30) & 12 \\
Rolling Cube & 76 & 52 & All-Angles-Bench (7); OmniSpatial (10); VisualTrans (35) & 12 \\
\addlinespace[2pt]
\rowcolor{black!5}\multicolumn{5}{l}{\textbf{C3: Constraint Reasoning}} \\
Hexakai & 64 & 0 & -- & 0 \\
Tapa & 75 & 10 & SSI-Bench (10) & 12 \\
Battleship & 44 & 20 & SSI-Bench (20) & 12 \\
Block Island & 67 & 70 & SSI-Bench (20); SpatiaLQA (50) & 12 \\
Pipe & 55 & 0 & -- & 0 \\
\addlinespace[2pt]
\rowcolor{black!5}\multicolumn{5}{l}{\textbf{C4: Sequential Planning}} \\
Laser & 47 & 0 & -- & 0 \\
KeyPath Quest & 64 & 29 & OmniSpatial (15); ReasonMap (14) & 12 \\
Growing Trail & 64 & 31 & VisualTrans (31) & 12 \\
Mancala & 55 & 40 & OmniSpatial (15); VisualTrans (25) & 12 \\
Rush Exit & 65 & 0 & -- & 0 \\
\midrule
\textbf{Total} & \textbf{1,000} & \textbf{400} & & \textbf{168} \\
\bottomrule
\end{tabularx}
\end{table}

\textbf{Model-level score pairs.}
Each of the 25 models in Table~\ref{tab:main_results} contributes one observation.
For model $m$, let $s_{m,i}$ denote the sample score returned by the corresponding
evaluator, and let $\mathcal{G}$ and $\mathcal{R}$ denote the complete toy-game
and real-world sets. The paired scores are
\begin{equation}
G_m=\frac{100}{1{,}000}\sum_{i\in\mathcal{G}}s_{m,i},
\qquad
R_m=\frac{100}{400}\sum_{i\in\mathcal{R}}s_{m,i}.
\label{g2r:eq:correlation_scores}
\end{equation}

Both are computed within their own evaluation set. In particular, neither
axis uses the joint score over all 1,400 samples. All 18 toy tasks contribute
to $G_m$, including those without a primary real-world alignment in
Table~\ref{g2r:tab:analysis_alignment}; the analysis is not restricted to the
14 tasks represented in the demonstration pool.

Writing the capability scores as $G_{m,c}$ and $R_{m,c}$, the same aggregation
can be expressed as
\begin{equation}
G_m=\frac{178G_{m,1}+222G_{m,2}+305G_{m,3}+295G_{m,4}}{1{,}000},
\qquad
R_m=\frac{1}{4}\sum_{c=1}^{4}R_{m,c}.
\label{g2r:eq:capability_weighted_scores}
\end{equation}
Thus, an unweighted mean of the four toy-game capability scores would produce
a different statistic, whereas the real-world mean is unchanged because its
capabilities have equal size. The observations are joined by model identity,
so each toy-game score remains paired with the real-world score of the same
model, irrespective of the display order in the main results table.

\textbf{Correlation coefficients.}
Let $M=25$, $\bar G=M^{-1}\sum_m G_m$, and
$\bar R=M^{-1}\sum_m R_m$. Pearson correlation is computed as
\begin{equation}
r=\frac{\sum_{m=1}^{M}(G_m-\bar G)(R_m-\bar R)}
{\sqrt{\sum_{m=1}^{M}(G_m-\bar G)^2}
 \sqrt{\sum_{m=1}^{M}(R_m-\bar R)^2}}.
\label{g2r:eq:pearson}
\end{equation}
Spearman correlation applies the same calculation to the ranks of the model
scores, with ranks determined separately within each evaluation set:
\begin{equation}
\rho=\operatorname{Corr}\!\left(
\operatorname{rank}(G_1,\ldots,G_M),
\operatorname{rank}(R_1,\ldots,R_M)\right).
\label{g2r:eq:spearman}
\end{equation}
Tied values receive average ranks. The implementation uses SciPy's
\texttt{pearsonr} for score correlation and \texttt{spearmanr} for rank
correlation. Applied to Table~\ref{g2r:tab:correlation_inputs}, these functions
give $r=0.9345988$ and $\rho=0.9346154$, both rounded to $0.935$ in the
main paper. The fitted line in Figure~\ref{fig:game_real_correlation}
visualizes the trend without entering either correlation calculation.

\begin{table}[H]
\centering
\caption{\textbf{The 25 paired scores used for cross-setting correlation.}
Values are the stored percentage scores used by the plotting script. Each
model appears once; the two blocks are a compact presentation of the same
25-observation analysis.}
\label{g2r:tab:correlation_inputs}
\small
\setlength{\tabcolsep}{3pt}
\renewcommand{\arraystretch}{1.08}
\begin{tabularx}{\linewidth}{@{}>{\raggedright\arraybackslash}X r r
@{\hspace{10pt}}>{\raggedright\arraybackslash}X r r@{}}
\toprule
\textbf{Model} & \textbf{Toy} & \textbf{Real} &
\textbf{Model} & \textbf{Toy} & \textbf{Real} \\
\midrule
GPT-5.6-Sol & 73.90 & 59.4750 & Gemma4-12B-it & 20.60 & 30.3300 \\
Gemini-3.1-Pro-Preview & 59.00 & 61.2750 & GLM-4.6V-Flash & 17.20 & 34.7400 \\
Claude-Opus-4.8-thinking & 55.00 & 50.4500 & Ovis2.5-9B & 17.00 & 28.2300 \\
Qwen3.8-27B & 50.30 & 58.2000 & InternVL3.5-8B & 15.10 & 28.0250 \\
Qwen3.5-27B & 42.40 & 51.1500 & Qwen3-VL-8B-Instruct & 13.90 & 28.1250 \\
Gemma4-31B-it & 34.70 & 47.8700 & MiMo-VL-7B-RL & 14.40 & 24.2750 \\
Qwen3.5-35B-A3B & 40.40 & 56.3000 & NEO-9B-SFT & 13.60 & 25.4000 \\
Qwen3-VL-30B-A3B-Instruct & 16.50 & 35.0250 & Qwen3.5-4B & 26.40 & 44.9750 \\
InternVL3.5-30B-A3B & 16.40 & 32.4000 & Qwen3.5-2B & 14.10 & 31.8750 \\
NEO-ov-9B & 15.30 & 27.9250 & NEO-2B-SFT & 9.50 & 22.2800 \\
SenseNova-U1-8B-MoT & 15.20 & 26.8250 & Qwen3-VL-2B-Instruct & 7.90 & 24.4500 \\
Qwen3.5-9B & 31.90 & 45.3250 & InternVL3.5-2B & 9.00 & 21.2000 \\
Step3-VL-10B & 20.60 & 33.1000 &  &  &  \\
\bottomrule
\end{tabularx}
\end{table}

The sample size is 25 models, not 1,400 questions or four capability averages.
The coefficients describe agreement in model-level performance across the
two settings, without establishing that the models use identical internal
reasoning processes to solve their different visual tasks.

\begingroup
\raggedbottom
\captionsetup[table]{font=normalsize,labelfont=bf,format=plain,
  justification=justified,singlelinecheck=false,margin=0pt,skip=5pt}
\subsection{Robustness of Model-Level Performance Correspondence}
\label{g2r:app:corr_robustness}

\textbf{Analysis scope and aggregation.}
We examine whether the model-level association between toy-game and real-world
performance persists under changes in capability weighting, model coverage,
and real-world task composition. These checks complement the primary
sample-weighted analysis. For model $m$, the four capability scores
$G_{m,c}$ and $R_{m,c}$ give the following averages:
\begin{equation}
    \widetilde G_m=\frac{1}{4}\sum_{c=1}^{4}G_{m,c},
    \qquad
    \widetilde R_m=\frac{1}{4}\sum_{c=1}^{4}R_{m,c}.
    \label{g2r:eq:robust_capability_scores}
\end{equation}
The real-world aggregate is unchanged because each capability contains 100
samples. On the toy-game side, equal capability weighting differs from the
sample-count weighting used in the primary analysis. Across the 25 models,
this alternative aggregation gives Pearson $r=0.935$ (95\% confidence interval:
$[0.903,0.974]$). Restricting the comparison to the 22 open-weight models gives
$r=0.960$ ($[0.913,0.981]$), so the strong association also appears within the
open-weight comparison. The analyses below retain the same $\widetilde G_m$
while varying the real-world aggregate, without modifying the benchmark's
primary scores or the predictions used to compute them.

\textbf{Real-world aggregation and scoring.}
For all 22 open-weight models, complete per-sample results reproduce the
real-world capability scores in Table~\ref{tab:main_results} within its
reported precision, with absolute differences below $0.046$ percentage points.
Table~\ref{g2r:tab:robust_real_aggregation} compares alternative aggregations.
Equal weighting of the 11 source means gives $r=0.964$. Excluding the 50
SpatiaLQA samples leaves 350 binary-scored samples, with $r=0.946$ under
sample averaging and $r=0.933$ under equal-capability averaging. Thus, the
observed correspondence remains strong both when source weights change and
when the continuous graph scores are excluded from the real-world aggregate.

\begin{table}[H]
\centering
\caption{\textbf{Sensitivity to real-world aggregation and answer format.}
All rows use the same 22 open-weight models and fixed toy-game
capability-balanced scores. Sample means weight retained samples equally;
capability means weight the four capability means equally. Source means
weight the 11 source means equally. Intervals use 10,000 paired model-bootstrap
replicates.}
\label{g2r:tab:robust_real_aggregation}
\small
\setlength{\tabcolsep}{4pt}
\renewcommand{\arraystretch}{1.08}
\begin{tabular*}{\linewidth}{@{\extracolsep{\fill}}lccc@{}}
\toprule
\textbf{Real-world aggregate or subset} & \textbf{Samples} & \textbf{Pearson $r$} & \textbf{95\% CI} \\
\midrule
All samples, sample mean & 400 & 0.960 & $[0.913,0.981]$ \\
All sources, source mean & 400 & 0.964 & $[0.902,0.987]$ \\
Without SpatiaLQA, sample mean & 350 & 0.946 & $[0.869,0.979]$ \\
Without SpatiaLQA, capability mean & 350 & 0.933 & $[0.837,0.972]$ \\
Single-choice only, sample mean & 236 & 0.939 & $[0.860,0.975]$ \\
Non-single-choice, sample mean & 164 & 0.857 & $[0.719,0.929]$ \\
Non-single-choice, capability mean & 164 & 0.766 & $[0.585,0.939]$ \\
\bottomrule
\end{tabular*}
\end{table}

\textbf{Answer format and retained coverage.}
The single-choice and non-single-choice subsets both retain positive
cross-setting associations, although the magnitudes differ. The 164
non-single-choice samples comprise 12 fine-grained perception, 38 spatial
reasoning, 100 constraint reasoning, and 14 sequential planning samples.
Sample and capability averaging therefore assign different weights to these
questions, yielding $r=0.857$ and $r=0.766$, respectively. Moreover, the retained
perception samples all come from GQA and the planning samples from ReasonMap.
This comparison changes the question mixture while keeping predictions and
scoring procedures fixed; it examines sensitivity to answer format and task
coverage, rather than comparing different evaluators on the same answers.

\textbf{Leave-one-source-out analysis.}
We remove each of the 11 real-world sources in turn and recompute the
correlation across the same 22 models. Table~\ref{g2r:tab:robust_source_removal}
reports both sample-weighted and capability-balanced real-world aggregates
for every removal. Pearson correlations range from $0.940$ to $0.966$ under
sample averaging and from $0.928$ to $0.966$ under capability averaging.
The association remains high after each individual source is removed,
showing that it does not depend on retaining any one source in this
real-world evaluation set under either aggregation rule.

\begin{table}[H]
\centering
\caption{\textbf{Leave-one-source-out correlations.}
Each row removes one real-world source while retaining all 22 models and
the same toy-game capability-balanced scores. The two right columns differ
only in how the remaining real-world samples are aggregated. The
retained predictions and scoring rules are unchanged, allowing comparison
under the two real-world aggregation rules.}
\label{g2r:tab:robust_source_removal}
\small
\setlength{\tabcolsep}{5pt}
\renewcommand{\arraystretch}{1.08}
\begin{tabular*}{\linewidth}{@{\extracolsep{\fill}}lccc@{}}
\toprule
\textbf{Removed source} & \textbf{Samples left} & \textbf{Sample mean $r$} & \textbf{Capability mean $r$} \\
\midrule
All-Angles-Bench & 380 & 0.961 & 0.963 \\
CAPTURe-real & 370 & 0.958 & 0.960 \\
GQA & 388 & 0.959 & 0.959 \\
IKEA Manuals at Work & 395 & 0.956 & 0.956 \\
OmniSpatial & 339 & 0.965 & 0.964 \\
ReasonMap & 386 & 0.959 & 0.958 \\
SPair-71k & 373 & 0.961 & 0.960 \\
SSI-Bench & 350 & 0.966 & 0.966 \\
SpatiaLQA & 350 & 0.946 & 0.933 \\
Visual-Jigsaw & 360 & 0.940 & 0.928 \\
VisualTrans & 309 & 0.949 & 0.953 \\
\bottomrule
\end{tabular*}
\end{table}

\textbf{Uncertainty estimates.}
The toy-game capability means use the one-decimal scores reported in
Table~\ref{tab:main_results}; real-world subset analyses use complete
per-sample results. Confidence intervals use 10,000 paired model-bootstrap
replicates with seed 20260925, keeping each model's score vector together
across the two settings. These intervals condition on the evaluated questions.
An additional bootstrap jointly resamples the 22 models and all 400
real-world samples, with sample resampling stratified by capability. It yields
an interval of $[0.881,0.981]$ over 5,000 replicates. This additional interval keeps the reported toy-game capability scores fixed
and includes variation from resampling real-world questions, but not
variation from resampling the independently generated toy-game questions.

\textbf{Relation to the main findings.}
Together, these checks support the stability of model-level score
correspondence across alternative weighting rules, model coverage, and
individual source removals. The answer-format results also show that its
magnitude varies with the retained task composition. These results describe
model-level correspondence, rather than the specificity of individual
capability matches. The one-shot experiment separately measures score changes after adding
randomly assigned solved game demonstrations to the unchanged real-world target questions.
\endgroup

\subsection{One-Shot Game-to-Real In-Context Learning}
\label{g2r:app:one_shot_icl}

\textbf{Demonstration pool.}
We study whether solved examples from a heterogeneous game pool provide
useful context for real-world visual reasoning. We use a fixed pool of
168 verified samples from Game2Real-Dataset, with 12 examples from each of
the 14 toy tasks represented by the real-world primary task annotations.
The pool contains 48 fine-grained perception, 48 spatial reasoning,
36 constraint reasoning, and 36 sequential planning examples
(Table~\ref{g2r:tab:analysis_alignment}). The game demonstrations remain
separate from the independently generated toy-game evaluation set. Each
example retains its task rules, visual observations, question, available
options, and verified final answer; intermediate solution traces are not
included in the demonstration presented to the model.

\textbf{Demonstration assignment.}
We use a common randomized policy to evaluate the shared pool without an
additional per-target retrieval or matching method. Because capability
labels describe primary but overlapping reasoning requirements, label
agreement is not used as a selection criterion; both same-capability and
cross-capability pairs are permitted. For each seed $s\in\{11,23,37\}$,
an assignment $a_s:\mathcal{R}\rightarrow\mathcal{D}$ selects one example
from the fixed pool $\mathcal{D}$ for every real-world target. Each of the
four demonstration capabilities is used in 100 of the 400 assignments.
Writing $c(d)$ for the capability of demonstration $d$, we impose the
following balance across the complete target set:
\begin{equation}
\sum_{i\in\mathcal{R}}\mathbf{1}\!\left[c(a_s(i))=c\right]=100,
\qquad c\in\{1,2,3,4\}.
\label{g2r:eq:demo_capability_balance}
\end{equation}
This global balance controls the overall demonstration mix despite unequal
capability-pool sizes, while the mix within each target capability is left
to the randomized assignment. It differs from sampling uniformly from all
168 examples. A demonstration may be reused across targets, while each
target receives only one demonstration within a particular assignment.

\textbf{Shared assignments across models.}
We evaluate Qwen3.5-2B, Qwen3.5-4B, Qwen3.5-9B, and Qwen3.8-27B on the same
400 real-world targets. Each seed specifies one complete demonstration--target
assignment, which is reused across all four models. The target images,
questions, answer conventions, and evaluators are unchanged across seeds;
the seeds vary the demonstration context rather than resampling the target
set. Model configurations are held fixed between the paired zero-shot and
one-shot evaluations, including the budgeted thinking setting when enabled.

\textbf{Role-aware prompt construction.}
Each one-shot input is a native conversation: a user turn supplies the game
images, rules, question, options when present, and answer instruction; an
assistant turn supplies its verified final answer; and a new user turn
supplies the real-world target. The target retains the \texttt{RealWorldFinal}
prompt from Appendix~\ref{g2r:app:prompting}. Role boundaries therefore
separate the solved demonstration from the question awaiting an answer.

\begin{figure}[H]
\centering
\begin{tcolorbox}[
  colback=black!2, colframe=black!40, boxrule=0.4pt,
  arc=1mm, left=1.5mm, right=1.5mm, top=1mm, bottom=1mm,
  fontupper=\small, title={One-shot conversation template},
  colbacktitle=black!8, coltitle=black, fonttitle=\small\bfseries]
\textbf{User turn: game demonstration}\par
\emph{Game images in their stored order}\par
\texttt{Rules:} \emph{Task rules, when present}\par
\emph{Game question}\par
\texttt{Options:} \emph{Answer options, when present}\par
\texttt{Give only the final answer in the requested format.}\par\smallskip
\textbf{Assistant turn: demonstrated answer}\par
\emph{Verified game final answer only}\par\smallskip
\textbf{User turn: real-world target}\par
\emph{Target images in their stored order}\par
\emph{Optional hint; target question; available options; image-role
instructions; task-specific answer-format instruction}
\end{tcolorbox}
\caption{\textbf{Conversation structure for one-shot evaluation.}
Italic text denotes sample-specific content, not additional literal
instructions. The demonstration's final-answer instruction is reproduced
verbatim; the target uses the unchanged benchmark prompt.}
\label{g2r:fig:one_shot_prompt}
\end{figure}

Image order is retained within each user turn, and the adapter's system
prompt is preserved. Target reference answers and evaluator metadata remain
outside the input supplied to the model.

\textbf{Averaging over assignments.}
Let $A^{(0)}_{m,c}$ denote the zero-shot score of model $m$ on real-world
capability $c$, and let $A^{(1,s)}_{m,c}$ denote the one-shot score under
assignment seed $s$. Here, $c$ groups the real-world targets, irrespective
of the capabilities of their assigned demonstrations. We average the three
evaluations and compare the resulting score with the same model's
zero-shot baseline:
\begin{equation}
\bar A^{(1)}_{m,c}
 =\frac{1}{3}\sum_{s\in\{11,23,37\}}A^{(1,s)}_{m,c},
\qquad
\Delta_{m,c}=\bar A^{(1)}_{m,c}-A^{(0)}_{m,c}.
\label{g2r:eq:icl_seed_average}
\end{equation}
Changes are expressed in percentage points. Overall scores average the four
equally sized real-world capabilities. The three assignments are averaged
at the score level, rather than combined by voting or best-response selection.
Cross-model capability gains likewise average the four model-specific changes,
with the same weight assigned to each model. Thus, $\Delta_{m,c}$ measures
the net effect of the randomized prompting policy on each target capability,
providing a baseline for future comparisons with capability-matched selection.
It reflects the full demonstration, including game content, formatting cues,
and added context, rather than separating their individual contributions.

\textbf{Results across demonstration assignments.}
Tables~\ref{g2r:tab:icl_seed_scores} and~\ref{g2r:tab:icl_seed_variation}
report all three assignments and their observed variation. The same 400
real-world targets and zero-shot baseline are used for each model;
only the assigned game demonstrations vary between the three evaluations.

\begin{table}[H]
\centering
\caption{\textbf{Zero-shot and per-assignment one-shot scores.}
C1--C4 denote fine-grained perception, spatial reasoning, constraint reasoning,
and sequential planning. Overall-400 averages the four equally sized target
groups. Scores are percentages. Bold rows report the three-assignment
means; the remaining rows show the zero-shot baseline or one-shot scores
for seeds 11, 23, and 37.}
\label{g2r:tab:icl_seed_scores}
\small
\setlength{\tabcolsep}{3pt}
\renewcommand{\arraystretch}{1.06}
\begin{tabular*}{\linewidth}{@{\extracolsep{\fill}}llccccc@{}}
\toprule
\textbf{Model} & \textbf{Setting} & \textbf{C1} & \textbf{C2} & \textbf{C3} & \textbf{C4} & \textbf{Overall-400} \\
\midrule
Qwen3.5-2B & Zero-shot & 41.00 & 23.00 & 16.51 & 47.00 & 31.88 \\
 & 1-shot seed 11 & 36.00 & 20.00 & 22.22 & 34.00 & 28.05 \\
 & 1-shot seed 23 & 38.00 & 23.00 & 22.08 & 41.00 & 31.02 \\
 & 1-shot seed 37 & 39.00 & 20.00 & 22.22 & 35.00 & 29.06 \\
 & \textbf{1-shot mean} & \textbf{37.67} & \textbf{21.00} & \textbf{22.17} & \textbf{36.67} & \textbf{29.38} \\
\midrule
Qwen3.5-4B & Zero-shot & 59.00 & 38.00 & 28.87 & 54.00 & 44.97 \\
 & 1-shot seed 11 & 41.00 & 42.00 & 21.21 & 52.00 & 39.05 \\
 & 1-shot seed 23 & 46.00 & 38.00 & 24.42 & 54.00 & 40.60 \\
 & 1-shot seed 37 & 50.00 & 40.00 & 27.84 & 55.00 & 43.21 \\
 & \textbf{1-shot mean} & \textbf{45.67} & \textbf{40.00} & \textbf{24.49} & \textbf{53.67} & \textbf{40.96} \\
\midrule
Qwen3.5-9B & Zero-shot & 51.00 & 52.00 & 24.32 & 54.00 & 45.33 \\
 & 1-shot seed 11 & 49.00 & 45.00 & 38.88 & 61.00 & 48.47 \\
 & 1-shot seed 23 & 54.00 & 36.00 & 40.96 & 60.00 & 47.74 \\
 & 1-shot seed 37 & 43.00 & 34.00 & 39.98 & 60.00 & 44.24 \\
 & \textbf{1-shot mean} & \textbf{48.67} & \textbf{38.33} & \textbf{39.94} & \textbf{60.33} & \textbf{46.82} \\
\midrule
Qwen3.8-27B & Zero-shot & 67.00 & 55.00 & 38.83 & 72.00 & 58.21 \\
 & 1-shot seed 11 & 61.00 & 57.00 & 41.27 & 68.00 & 56.82 \\
 & 1-shot seed 23 & 70.00 & 53.00 & 42.15 & 68.00 & 58.29 \\
 & 1-shot seed 37 & 68.00 & 55.00 & 41.08 & 70.00 & 58.52 \\
 & \textbf{1-shot mean} & \textbf{66.33} & \textbf{55.00} & \textbf{41.50} & \textbf{68.67} & \textbf{57.88} \\
\bottomrule
\end{tabular*}
\end{table}

\begin{table}[H]
\centering
\caption{\textbf{Overall changes and assignment variability.}
$\Delta$ is relative to the same model's zero-shot score. SD is the sample
standard deviation across the three one-shot scores. Changes and SD are in
percentage points. Ranges are observed minima and maxima, not confidence
intervals; the last column counts how many of the three assignments improve
over the zero-shot baseline.}
\label{g2r:tab:icl_seed_variation}
\small
\setlength{\tabcolsep}{4pt}
\renewcommand{\arraystretch}{1.06}
\begin{tabular*}{\linewidth}{@{\extracolsep{\fill}}lcccc@{}}
\toprule
\textbf{Model} & \textbf{Mean $\Delta$} & \textbf{SD} & \textbf{Observed $\Delta$ range} & \textbf{Gains / seeds} \\
\midrule
Qwen3.5-2B & $-2.50$ & 1.51 & $[-3.83,-0.86]$ & 0/3 \\
Qwen3.5-4B & $-4.01$ & 2.10 & $[-5.92,-1.76]$ & 0/3 \\
Qwen3.5-9B & $+1.49$ & 2.26 & $[-1.09,+3.14]$ & 2/3 \\
Qwen3.8-27B & $-0.33$ & 0.92 & $[-1.39,+0.31]$ & 2/3 \\
\bottomrule
\end{tabular*}
\end{table}

Constraint reasoning improves under all three assignments for the 2B, 9B,
and 27B models, but declines for the 4B model. Qwen3.5-9B gains $14.56$--$16.64$
points in this capability. Averaging across the four models separately for
each assignment gives constraint-reasoning gains of $3.76$, $5.27$, and $5.65$
points for seeds 11, 23, and 37. For each assignment, the corresponding
four-model averages for fine-grained perception and Overall-400 remain
below their zero-shot baselines.

Overall changes are negative under all three assignments for the 2B and 4B
models. The 9B and 27B models each improve in two assignments, although their
mean changes have opposite signs ($+1.49$ and $-0.33$ points, respectively).
For the 27B model, the small overall mean combines a constraint-reasoning
gain ($+2.67$) with a planning decline ($-3.33$). These summaries describe
variation across demonstration assignments, with all 400 target questions
held fixed in every evaluation.

\subsection{Thinking-Mode Configuration}
\label{g2r:app:thinking_configuration}

\textbf{Evaluation scope.}
We compare the main-evaluation and non-thinking configurations of Qwen3.5-2B,
Step3-VL-10B, Gemma4-31B-it, and GPT-5.6-Sol on all 1,400 benchmark samples.
Both configurations use the same questions, images, answer formats, and
evaluators in zero-shot evaluation, without game demonstrations.

\textbf{Inference settings.}
Qwen3.5-2B follows the conditional $32\mathrm{k}+512$-token thinking protocol
in Appendix~\ref{g2r:app:model_configurations}. For GPT-5.6-Sol, the non-thinking
configuration sets \texttt{reasoning\_effort="none"}, with zero reasoning tokens
in the usage records; the main evaluation leaves this field unspecified and
uses the API default. Their output limits are 32,768 and 16,384 tokens,
respectively, with no additional answer-completion stage.
For Gemma4-31B-it, the non-thinking configuration sets
\texttt{enable\_thinking=False}, with greedy decoding and a 4,096-token output
limit. The main run leaves the flag unspecified; its recorded wrapper default
is also \texttt{False}. For Step3-VL-10B, the non-thinking template prefills an
empty \texttt{<think></think>} block, with greedy decoding and a 16,384-token
output limit. The main evaluation passes no suppression flag and uses
temperature $1.0$, top-$p=0.95$, top-$k=20$, and a 40,960-token output limit;
its server-side template was not retained.

\textbf{Score reporting.}
Table~\ref{g2r:tab:thinking_comparison} reports the mean per-sample score over
the full benchmark and the gain in percentage points relative to the
non-thinking configuration. The gains summarize the complete inference
configurations, including their decoding settings and output budgets.

\begin{table}[H]
\centering
\caption{\textbf{Overall scores under different inference settings.}
Both columns use the full 1,400-sample benchmark. Main evaluation reports the
scores from the main experiments; Gain is its difference from Non-thinking
in percentage points.}
\label{g2r:tab:thinking_comparison}
\small
\setlength{\tabcolsep}{6pt}
\renewcommand{\arraystretch}{1.08}
\begin{tabular*}{\linewidth}{@{\extracolsep{\fill}}lrrr@{}}
\toprule
\textbf{Model}
& \textbf{Non-thinking}
& \textbf{Main evaluation}
& \textbf{Gain} \\
\midrule
Qwen3.5-2B
& 13.07 & 19.30 & +6.23 \\
Step3-VL-10B
& 18.25 & 24.20 & +5.95 \\
Gemma4-31B-it
& 27.41 & 38.50 & +11.09 \\
GPT-5.6-Sol
& 28.27 & 69.80 & +41.53 \\
\bottomrule
\end{tabular*}
\end{table}

\end{document}